\documentclass[10pt, conference, compsocconf]{IEEEtran}

\usepackage[hscale=0.8,vscale=0.8]{geometry}
\usepackage{amsmath}
\usepackage{graphicx}
\usepackage{amssymb}
\usepackage{algorithm}
\usepackage{algpseudocode}
\usepackage{textcomp}
\usepackage{listings}
\usepackage[usenames,dvipsnames]{xcolor}
\usepackage{subcaption}
\usepackage{cite}
\usepackage{hyperref}
\usepackage{lipsum}
\usepackage[normalem]{ulem}
\usepackage{listings} 
\usepackage{paralist}
\usepackage{algpseudocode}
\usepackage{url}
\usepackage{array}
\usepackage{bm}
\usepackage[justification=centering]{caption}
\usepackage{tikz}
\usetikzlibrary{positioning, fit, arrows.meta, shapes}
\allowdisplaybreaks

\newcommand\tab[1][0.5cm]{\hspace*{#1}}

\begin{document}
\title{Learning Deep Representations from Clinical Data for Chronic Kidney Disease}

\author{\IEEEauthorblockN{Duc Thanh Anh Luong}
\IEEEauthorblockA{Department of Computer Science and Engineering\\
University at Buffalo\\
Buffalo, New York 14260\\
Email: ducthanh@buffalo.edu}
\and
\IEEEauthorblockN{Varun Chandola}
\IEEEauthorblockA{Department of Computer Science and Engineering\\
University at Buffalo\\
Buffalo, New York 14260\\
Email: chandola@buffalo.edu}}

\maketitle


\begin{abstract} \small\baselineskip=9pt
We study the behavior of a Time-Aware Long Short-Term Memory Autoencoder, a state-of-the-art method, in the context of learning latent representations from irregularly sampled patient data. We identify a key issue in the way such recurrent neural network models are being currently used and show that the solution of the issue leads to significant improvements in the learnt representations on both synthetic and real datasets. A detailed analysis of the improved methodology for representing patients suffering from Chronic Kidney Disease (CKD) using clinical data is provided. Experimental results show that the proposed T-LSTM model is able to capture the long-term trends in the data, while effectively handling the noise in the signal. Finally, we show that by using the latent representations of the CKD patients obtained from the T-LSTM autoencoder, one can identify unusual patient profiles from the target population.
\end{abstract}

\section{Introduction}
Chronic Kidney Disease (CKD) is a world-wide public health problem, which was responsible for approximately 956,200 deaths globally in 2013~\cite{abubakar2015global}. In CKD, the primary indicator of disease severity is \textit{estimated glomerular filtration rate} (eGFR) - a clinical marker that measures kidney function.
Although CKD is a chronic condition in which kidney function deteriorates over time, the courses of disease progressions among patients are highly heterogeneous. From a clinical perspective, being able to identify subgroups of patients with similar CKD disease progressions is a first step toward disease subtyping as discovered patient sub-groups may exhibit common biological or physiological patterns that allow us to have better understanding of the underlying mechanism of disease within sub-groups and further develop better treatments for those patients. 

From a machine learning perspective, the task of identifying groups of patients with similar disease progressions is an unsupervised machine learning problem in which patient progressions exhibited by their eGFR values over time are clustered. Most recent solutions, in the context of CKD, have relied on fitting a set of functions, e.g., splines, polynomials, etc., to represent the disease trajectories~\cite{luong2017egems,luong2017kmeans,singh2017automatic}. However, such models require pre-specifying the form and number of such functions, which becomes a limiting factor. At the same time, recurrent neural network architectures, such as {\em Long Short-Term Memory networks} or LSTMs, have become the workhorses for modeling sequence data. Recently, LSTMs have been applied to clinical data for various analytical tasks~\cite{baytas2017patient,WU2018167}. 
To handle the issue of irregular sampling, which is typical in clinical settings, a recent modification called Time-Aware LSTM (T-LSTM) has been proposed~\cite{baytas2017patient}. 
T-LSTMs have been shown to capture short and long-range temporal dependencies in the data and learn latent representations from irregularly sampled observations for patients suffering from Parkinson disease. 

While the LSTM-based latent representation learning methodology has been studied in limited disease settings, e.g., Parkinson's disease~\cite{baytas2017patient} and Asthma~\cite{WU2018167}, it is unclear how the same architecture would perform in a different disease context, i.e., CKD. In fact, LSTMs offer more than one way of inferring the latent representation for a given input profile. Current applications have used one of those --- the activations at the hidden unit (See Section~\ref{sec:experiment_synthetic} for more details). Will the same methodology be applicable for CKD? At a more fundamental level, the relationship between the activations within different parts of the network and the properties of the input temporal sequence, remains unstudied. 

In this paper we explore the above-mentioned relationship to devise a better strategy for learning latent representation of patient disease progression profiles. Through results on synthetically generated temporal sequences, we discover a better representation and then use the improved representation to understand CKD progression. 

\subsection*{Key contributions}
\begin{inparaenum}[1)]
  \item We study the behavior of T-LSTM autoencoder using synthetically generated datasets and show that a latent representation that uses activations at hidden and memory units is better than using the hidden unit activation alone, which is the prevalent strategy.
  \item An in-depth analysis of the T-LSTM autoencoder, in the context of CKD, using two real-world datasets, is provided, which includes a data-driven strategy to choose key network parameters.
  \item An application of the improved solution to identify anomalous patient profiles is demonstrated.
\end{inparaenum}

The remaining of our paper is structured as follows. In Section~\ref{sec:method}, we provide a brief background on T-LSTM Autoencoder, the primary method that we use in this paper. In Section~\ref{sec:experiment_synthetic}, we present our approach to represent the longitudinal profiles in the embedded space by using three different synthetic datasets. In Section~\ref{sec:experiment}, we present experimental results with T-LSTM Auto-encoder on CKD datasets. In Section~\ref{sec:discussion}, we discuss the advantages and drawbacks of this method when applying on CKD as well as other related works. Finally, in Section~\ref{sec:conclusion}, we give a conclusion of our paper. 

\section{Background}
\label{sec:method}
This section provides necessary background to understand the Time-Aware LSTM autoencoder and its suitability for handling longitudinal clinical data.
\subsection{Recurrent Neural Networks (RNN)}
A typical input sequence for a RNN is of the form $\{{\bf x}_t\}_{t=1}^T$ where $T$ is the length of the input sequence and ${\bf x}_t$ is a vector of size $N$. In vanilla RNN, at each time step $t$, there is a hidden unit ${\bf h}_t \in \mathbb{R}^{H}$ computed by combining both the input signal at this time step ${\bf x}_t$ and the hidden unit of previous time step ${\bf h}_{t-1}$: 
\begin{equation}
	{\bf h}_t = \tanh\left({\bf W} {\bf x}_t + {\bf U} {\bf h}_{t-1} + b\right)\nonumber
	\label{eq:rnn_h}
\end{equation}
With this network construction, the hidden unit ${\bf h}_t$ plays a role of memory that captures the information of all previous inputs $\{{\bf x}_i\}_{i=1}^t$. 
\subsection{Long Short-Term Memory (LSTM) Networks}
For RNN, learning the network parameters is typically done by applying traditional learning algorithm such as gradient descent with back-propogation through time. However, for applications with very long time lags (many steps between the signal and the output), the learning algorithm may suffer from the problem of exploding or vanishing gradients.
This problem is addressed in LSTM architecture~\cite{hochreiter1997long} by enforcing a ``constant error carousel" within special units, called {\em memory units}. 
The information stored in these units is controlled by a gated structure. 

Typically, at a time step $t$, beside the hidden unit ${\bf h}_t$, LSTM has dedicated memory cells ${\bf C}_t$. The information in memory cells ${\bf C}_t$ is controlled by ``forget gate" ${\bf f}_t$ and ``input gate" ${\bf i}_t$. 
The information from previous memory cells can be moved forward to be stored again in the current memory cells, depending on the values of previous hidden units and current inputs.
\subsection{Time-Aware LSTM Networks}
\label{subsec:tlstm}

A key assumption with the LSTM model (and also vanilla RNN) is that temporal sequence is regularly sampled. 
If the sequence has few missing values, one can extend the dimension of input ${\bf x}_t$ to encode whether there is missing values in the input. 
However, for clinical datasets such as patient's lab test results, the time gaps between consecutive observations can be multiple days or weeks or even months. 
This makes the approach of encoding missing values within the input ${\bf x}_t$ infeasible as there are too many missing values in comparison with actual observations. 
In addition, the time gaps between observations do carry the signal in themselves. 
For example, in clinical applications, having fewer lab tests within a certain time span may indicate that a patient has good health and therefore does not require to take lab tests regularly.
In contrast, having lab tests more often within a certain period may indicate that a patient has some clinical problems and doctor needs to monitor her physiological markers more frequently.
With the lack of proper representation of time lapse between consecutive observations in traditional RNN models, Time-Aware LSTM (T-LSTM) is proposed by Baytas et al.~\cite{baytas2017patient} to address this issue. 
Figure~\ref{fig:tlstm} shows the structure of a T-LSTM unit. 

In T-LSTM, the previous memory cells ${\bf C}_t$ is decomposed into short-term memory ${\bf C}^S_t$ and long-term memory ${\bf C}^T_t$. Only the short-term memory interacts with the time gap $\Delta_t$ to adjust for the time lapse since previous observation. In particular, the non-increasing function $g(\dot)$ is first applied on $\Delta_t$ to penalize for the time passed since previous observation. The quantity $g(\Delta_t)$ now becomes a multiplier to adjust for short-term memory ${\bf C}^S_t$. The longer time passed since previous observation, the less amount of short-term memory is kept to use in the next calculation.

Beside the decomposition of previous memory cells and the interaction between short-term memory and time lapse $\Delta_t$, the structure of T-LSTM is exactly the same with LSTM. 
The detailed formulation of T-LSTM is shown as follows:
  \begin{align}
    {\bf C}^S_{t-1} &=&& \tanh\left({\bf W}_d {\bf C}_{t-1} + {\bf b}_d \right) \text{\tab(short-term memory)}& \nonumber\\
    {\bf \hat{C}}^S_{t-1} &=&& {\bf C}^S_{t-1} * g(\Delta_t) \text{\tab(discounted short-term memory)}& \nonumber\\
    {\bf C}^T_{t-1} &=&& {\bf C}_{t-1} - {\bf C}^S_{t-1} \text{\tab(long-term memory)}& \nonumber\\
    {\bf C}^*_{t-1} &=&& {\bf C}^T_{t-1} + {\bf \hat{C}}^S_{t-1} \text{\tab(adjusted previous memory)}& \nonumber\\
    {\bf f}_t &=&& \sigma\left({\bf W}_f {\bf x}_t + {\bf U}_f {\bf h}_{t-1} + {\bf b}_f \right) \text{\tab(forget gate)}&\nonumber\\
    {\bf i}_t &=&& \sigma\left({\bf W}_i {\bf x}_t + {\bf U}_i {\bf h}_{t-1} + {\bf b}_i \right) \text{\tab(input gate)}&\nonumber\\
    {\bf \tilde{C}}_t &=&& \tanh\left({\bf W}_c {\bf x}_t + {\bf U}_c {\bf h}_{t-1} + {\bf b}_c \right) \text{\tab(candidate memory)}&\nonumber\\
    {\bf C}_t &=&& {\bf f}_t * {\bf C}^*_{t-1} + {\bf i}_t * {\bf \tilde{C}}_t \text{\tab(current memory)}&\nonumber\\
    {\bf o}_t &=&& \sigma\left({\bf W}_o {\bf x}_t + {\bf U}_o {\bf h}_{t-1} + {\bf b}_o \right) \text{\tab(output gate)}&\nonumber\\	
    {\bf h}_t &=&& {\bf o}_t * \tanh({\bf C}_t) \text{\tab(current hidden state)}& \nonumber
  \end{align}
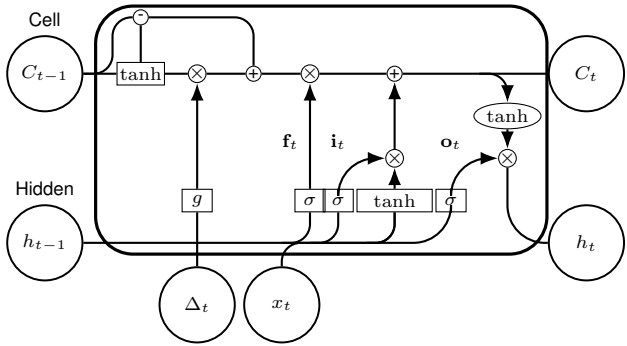
\begin{figure}[!ht]
  \centering
  \begin{tikzpicture}[
    scale=0.75,
    font=\sf \scriptsize,
    >=LaTeX,
    cell/.style={
        rectangle, 
        rounded corners=5mm, 
        draw,
        very thick,
        },
    operator/.style={
        circle,
        draw,
        inner sep=-0.5pt,
        minimum height =.2cm,
        },
    function/.style={
        ellipse,
        draw,
        inner sep=1pt
        },
    ct/.style={
        circle,
        draw,
        line width = .75pt,
        minimum width=1cm,
        inner sep=1pt,
        },
    gt/.style={
        rectangle,
        draw,
        minimum width=4mm,
        minimum height=3mm,
        inner sep=1pt
        },
    mylabel/.style={
        font=\scriptsize\sffamily
        },
    ArrowC1/.style={
        rounded corners=.25cm,
        thick,
        },
    ArrowC2/.style={
        rounded corners=.5cm,
        thick,
        },
    ]

    \node [cell, minimum height =3.3cm, minimum width=6.0cm] at (-1.8,0.5){} ;

	\node [gt] (tanh_c) at (-5,1.5) {$\tanh$};
	\node [operator] (mul_c) at (-4,1.5) {$\times$};
	\node [operator] (add_c) at (-3,1.5) {+};
	\node [operator] (sub_c) at (-5,2.5) {-};	
	
    \node [gt, label={[label distance=0.4cm, font=\scriptsize\sffamily]93:${\bf f}_t$}] (sigma_f) at (-2,-0.75) {$\sigma$};
    \node [gt, label={[label distance=0.4cm, font=\scriptsize\sffamily]90:${\bf i}_t$}] (sigma_i) at (-1.5,-0.75) {$\sigma$};
    \node [gt, minimum width=1cm] (tanh_i) at (-0.5,-0.75) {$\tanh$};
    \node [gt, label={[label distance=0.4cm, font=\scriptsize\sffamily]90:${\bf o}_t$}] (sigma_o) at (0.5,-0.75) {$\sigma$};
	\node [gt] (g) at (-4, -0.75) {$g$};	
	
    \node [operator] (mul_f) at (-2,1.5) {$\times$};
    \node [operator] (add_memory_update) at (-0.5,1.5) {+};
    \node [operator] (mul_i) at (-0.5,0) {$\times$};
    \node [operator] (mul_o) at (1.5,0) {$\times$};
    \node [function] (tanh_o) at (1.5,0.75) {$\tanh$};

    \node[ct, label={[mylabel]Cell}] (c_prev) at (-6.7,1.5) {$C_{t-1}$};
    \node[ct, label={[mylabel]Hidden}] (h_prev) at (-6.7,-1.5) {$h_{t-1}$};
    \node[ct] (x) at (-2.5,-2.6) {$x_t$};
	\node[ct] (delta_t) at (-4,-2.6) {$\Delta_t$};

    \node[ct] (c) at (2.9,1.5) {$C_t$};

    \node[ct] (h) at (2.9,-1.5) {$h_t$};

    \draw [ArrowC1] (c_prev) -- (tanh_c) -- (mul_c) -- (add_c) -- (mul_f) -- (add_memory_update) -- (c);

    \draw [ArrowC2] (h_prev) -| (sigma_o);
    \draw [ArrowC1] (h_prev -| sigma_f)++(-0.5,0) -| (sigma_f); 
    \draw [ArrowC1] (h_prev -| sigma_i)++(-0.5,0) -| (sigma_i);
    \draw [ArrowC1] (h_prev -| tanh_i)++(-0.5,0) -| (tanh_i);
    \draw [ArrowC1] (x) -- (x |- h_prev)-| (tanh_i);

    \draw [->, ArrowC2] (sigma_f) -- (mul_f);
    \draw [->, ArrowC2] (sigma_i) |- (mul_i);
    \draw [->, ArrowC2] (tanh_i) -- (mul_i);
    \draw [->, ArrowC2] (sigma_o) |- (mul_o);
    \draw [->, ArrowC2] (mul_i) -- (add_memory_update);
    \draw [->, ArrowC1] (add_memory_update -| tanh_o)++(-0.5,0) -| (tanh_o);
    \draw [->, ArrowC2] (tanh_o) -- (mul_o);

    \draw [-, ArrowC2] (mul_o) |- (h);
	
	\draw [-, ArrowC2] (delta_t) -- (g);
	\draw [->, ArrowC2] (g) -- (mul_c); 
	\draw [-, ArrowC1] (c_prev) -- ++(1.2,0) -- ++(0,1) -- (sub_c) -- ++(2,0) -- (add_c);
	\draw [-, ArrowC1] (tanh_c) -- (sub_c);
	
\end{tikzpicture}
  \caption{Illustration of a T-LSTM unit}
  \label{fig:tlstm}
\end{figure}

\subsection{T-LSTM Autoencoder}
\label{subsec:autoencoder}

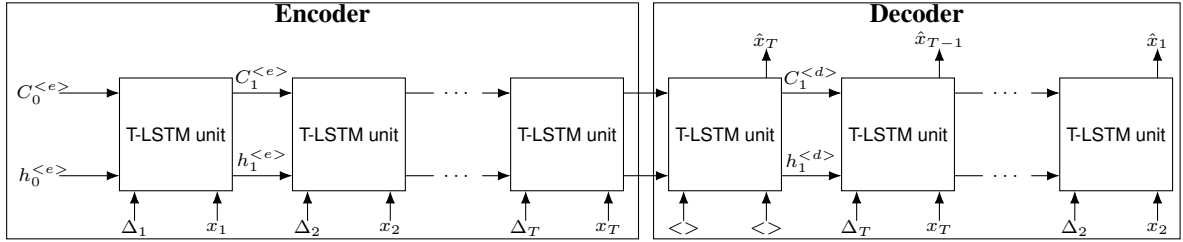
\begin{figure*}
  \centering
  \begin{tikzpicture}[
    font=\sf \scriptsize,
    >=LaTeX,
    cell/.style={
        rectangle, 
        rounded corners=5mm, 
        draw,
        very thick,
        },
    operator/.style={
        circle,
        draw,
        inner sep=-0.5pt,
        minimum height =.2cm,
        },
    function/.style={
        ellipse,
        draw,
        inner sep=1pt
        },
    ct/.style={
        circle,
        draw,
        line width = .75pt,
        minimum width=1cm,
        inner sep=1pt,
        },
    gt/.style={
        rectangle,
        draw,
        minimum width=4mm,
        minimum height=3mm,
        inner sep=1pt
        },
    mylabel/.style={
        font=\scriptsize\sffamily
        },
    ArrowC1/.style={
        rounded corners=.25cm,
        thick,
        },
    ArrowC2/.style={
        rounded corners=.5cm,
        thick,
        },
    ]

	\draw (-1.5, -0.64) rectangle (6.9, 2.5);
	\draw (7.1, -0.64) rectangle (14.1, 2.5);	
	\node [font=\bfseries] at (2.7, 2.36) {Encoder};
	\node [font=\bfseries] at (10.6, 2.36) {Decoder};
	
	\draw (0,0) rectangle (1.5,1.5) node[pos=.5] {T-LSTM unit};   
	\draw (2.3,0) rectangle (3.8,1.5) node[pos=.5] {T-LSTM unit};  
	\draw (5.2,0) rectangle (6.7,1.5) node[pos=.5] {T-LSTM unit};  
	\draw (7.3,0) rectangle (8.8,1.5) node[pos=.5] {T-LSTM unit};  
	\draw (9.6,0) rectangle (11.1,1.5) node[pos=.5] {T-LSTM unit};  
	\draw (12.5,0) rectangle (14,1.5) node[pos=.5] {T-LSTM unit};  
	
	\draw [->] (0.2,-0.4) -- (0.2,0);
	\node [mylabel] at (0.2,-0.5) {$\Delta_1$};
	\draw [->] (1.3,-0.4) -- (1.3,0);
	\node [mylabel] at (1.3,-0.5) {$x_1$};
	
	\draw [->] (2.5,-0.4) -- (2.5,0);
	\node [mylabel] at (2.5,-0.5) {$\Delta_2$};
	\draw [->] (3.6,-0.4) -- (3.6,0);
	\node [mylabel] at (3.6,-0.5) {$x_2$};	
		
	\draw [->] (5.4,-0.4) -- (5.4,0);
	\node [mylabel] at (5.4,-0.5) {$\Delta_T$};
	\draw [->] (6.5,-0.4) -- (6.5,0);
	\node [mylabel] at (6.5,-0.5) {$x_T$};	
	
    \draw [->] (7.5,-0.4) -- (7.5,0);
	\node [mylabel] at (7.5,-0.5) {$<>$};
	\draw [->] (8.6,-0.4) -- (8.6,0);
	\node [mylabel] at (8.6,-0.5) {$<>$};
	
	\draw [->] (9.8,-0.4) -- (9.8,0);
	\node [mylabel] at (9.8,-0.5) {$\Delta_T$};
	\draw [->] (10.9,-0.4) -- (10.9,0);
	\node [mylabel] at (10.9,-0.5) {$x_T$};
	
	\draw [->] (12.7,-0.4) -- (12.7,0);
	\node [mylabel] at (12.7,-0.5) {$\Delta_2$};
	\draw [->] (13.8,-0.4) -- (13.8,0);
	\node [mylabel] at (13.8,-0.5) {$x_2$};
    
	\draw [->] (8.6, 1.5) -- (8.6, 1.9);
	\node [mylabel] at (8.6, 2.0) {$\hat{x}_T$};
	\draw [->] (10.9, 1.5) -- (10.9, 1.9);
	\node [mylabel] at (10.9, 2.0) {$\hat{x}_{T-1}$};
	\draw [->] (13.8, 1.5) -- (13.8, 1.9);
	\node [mylabel] at (13.8, 2.0) {$\hat{x}_{1}$};

	\node [mylabel] (dots_lower_encoder) at (4.5,0.2) {$\dots$};
	\node [mylabel] (dots_upper_encoder) at (4.5,1.3) {$\dots$};
	\node [mylabel] (dots_lower_decoder) at (11.8,0.2) {$\dots$};
	\node [mylabel] (dots_upper_decoder) at (11.8,1.3) {$\dots$};
	
	\node [mylabel] (h_0) at (-1,0.2) {$h^{<e>}_0$};
	\node [mylabel] (C_0) at (-1,1.3) {$C^{<e>}_0$};
	\node [mylabel] (h_1) at (1.9,0.4) {$h^{<e>}_1$};
	\node [mylabel] (C_1) at (1.9,1.5) {$C^{<e>}_1$};
	
	\node [mylabel] (h_1_d) at (9.2,0.4) {$h^{<d>}_1$};
	\node [mylabel] (C_1_d) at (9.2,1.5) {$C^{<d>}_1$};
	
	\draw [->] (-0.8, 0.2) -- (0, 0.2);
	\draw [->] (-0.8, 1.3) -- (0, 1.3);
	\draw [->] (1.5,0.2) -- (2.3, 0.2);
	\draw [->] (1.5,1.3) -- (2.3, 1.3);
	\draw [->] (3.8,0.2) -- (dots_lower_encoder) -- (5.2,0.2);
    \draw [->] (3.8,1.3) -- (dots_upper_encoder) -- (5.2, 1.3);
	\draw [->] (6.7,0.2) -- (7.3, 0.2);
	\draw [->] (6.7,1.3) -- (7.3, 1.3);
	\draw [->] (8.8,0.2) -- (9.6, 0.2);
	\draw [->] (8.8,1.3) -- (9.6, 1.3);
	\draw [->] (11.1,0.2) -- (dots_lower_decoder) -- (12.5, 0.2);
	\draw [->] (11.1,1.3) -- (dots_upper_decoder) -- (12.5, 1.3);
\end{tikzpicture}
  \caption{T-LSTM Autoencoder}
  \label{fig:tlstm_ae}
\end{figure*}

With T-LSTM unit as a building block, it is possible to map complex patient longitudinal profile into an embedded space by using an autoencoder structure and analyze disease stratification in this embedded space. In Baytas et al.'s study~\cite{baytas2017patient}, T-LSTM Autoencoder is represented by two components including one encoder and one decoder. The illustration of T-LSTM autoencoder is shown in Figure~\ref{fig:tlstm_ae}. In this figure, the structure of encoder and decoder are exactly the same except that the T-LSTM unit in the decoder has another set of learnable parameters ${\bf W}^{<d>}_{out}$ and ${\bf b}^{<d>}_{out}$ to produce outputs from the hidden unit:
\begin{equation}
  \hat{\bf x}_t = {\bf W}^{<d>}_{out} {\bf h}^{<d>}_t + {\bf b}^{<d>}_{out}\nonumber
\end{equation}
In formulation of T-LSTM Autoencoder, we use the superscript $<e>$ and $<d>$ to distinguish the parameters of encoder and decoder respectively. In T-LSTM Autoencoder, the hidden unit and memory unit output from encoder are subsequently passed into the decoder so that it can reconstruct the original longitudinal observations in reverse chronological order. In particular, each T-LSTM unit in the decoder receives two inputs - (1) time gap from previous observation $\Delta_{t}$ and (2) the current observation ${\bf x}_t$; and outputs the prediction of previous observation $\hat{\bf x}_{t-1}$. For the first time step in the decoder, we define the current observation is a zero vector and time gap from previous observation is zero. The difference between the output prediction $\hat{\bf x}_t$ and actual observation ${\bf x}_t$ is the training signal that gradient-based learning algorithm can use to back-propagate the error through the decoder and encoder and learn to update the model's parameters.

\section{Representation of Longitudinal Profiles in T-LSTM Auto-encoder}
\label{sec:experiment_synthetic}
As illustrated in Figure~\ref{fig:tlstm_ae}, the hidden unit ${\bf h}^{<e>}_T$ and memory unit ${\bf C}^{<e>}_T$ at the last time step of the encoder will be fed into the decoder as the initial hidden unit and memory unit, i.e. ${\bf h}^{<d>}_0$ and ${\bf C}^{<d>}_0$. 
Although both hidden unit ${\bf h}^{<e>}_T$ and memory unit ${\bf C}^{<e>}_T$ are passed from encoder to decoder, in the original paper of Baytas et al.~\cite{baytas2017patient}, only the hidden unit ${\bf h}^{<e>}_T$ is used as the representation of the input longitudinal profile. 
This raises a question whether the hidden unit alone is sufficient to represent the longitudinal profile and whether the memory unit should be part of the representation. 
In this section, we investigate different possible representations of a longitudinal profile using T-LSTM Autoencoder and determine the most probable representation.

\begin{figure}[!ht]
	\begin{subfigure}[c]{0.5\textwidth}
		\includegraphics[trim={0.45cm 0 0.45cm 0.3cm},clip, width=28mm, angle=-90]{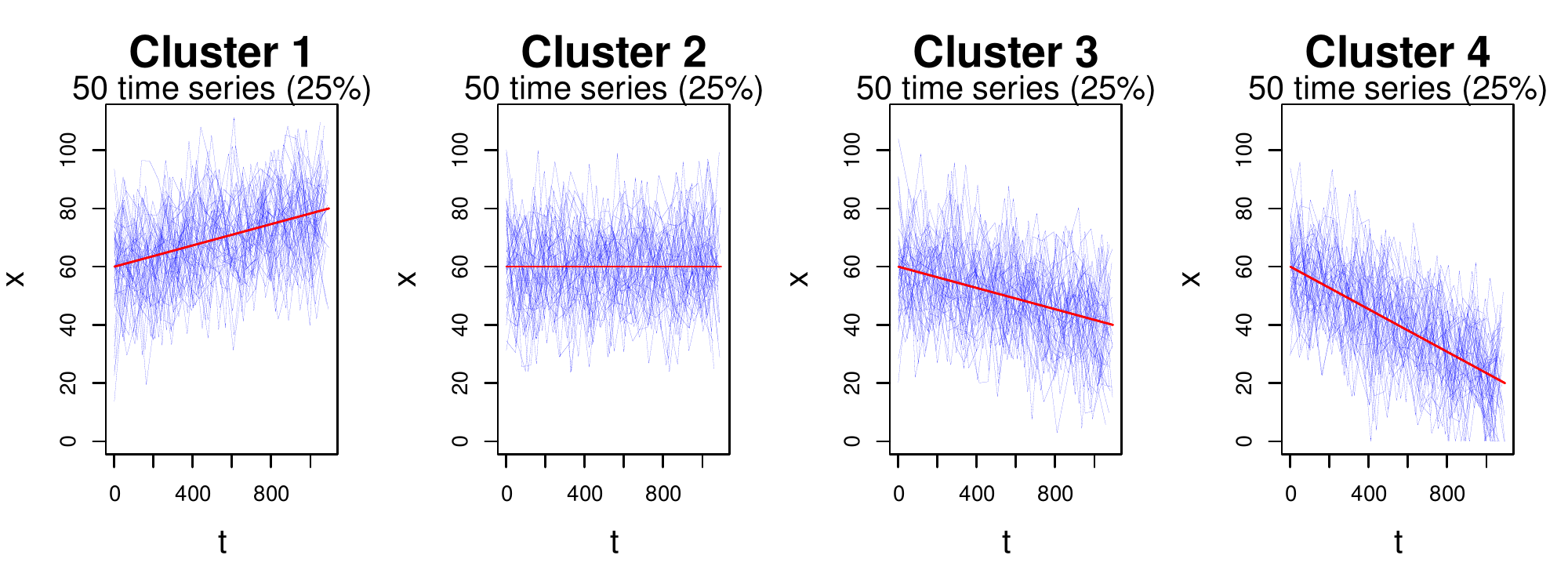}
		\caption{Synthetic dataset 1}
	\end{subfigure}
	\begin{subfigure}[c]{0.5\textwidth}
		\includegraphics[trim={0.45cm 0 0.45cm 0.3cm},clip, width=28mm, angle =-90]{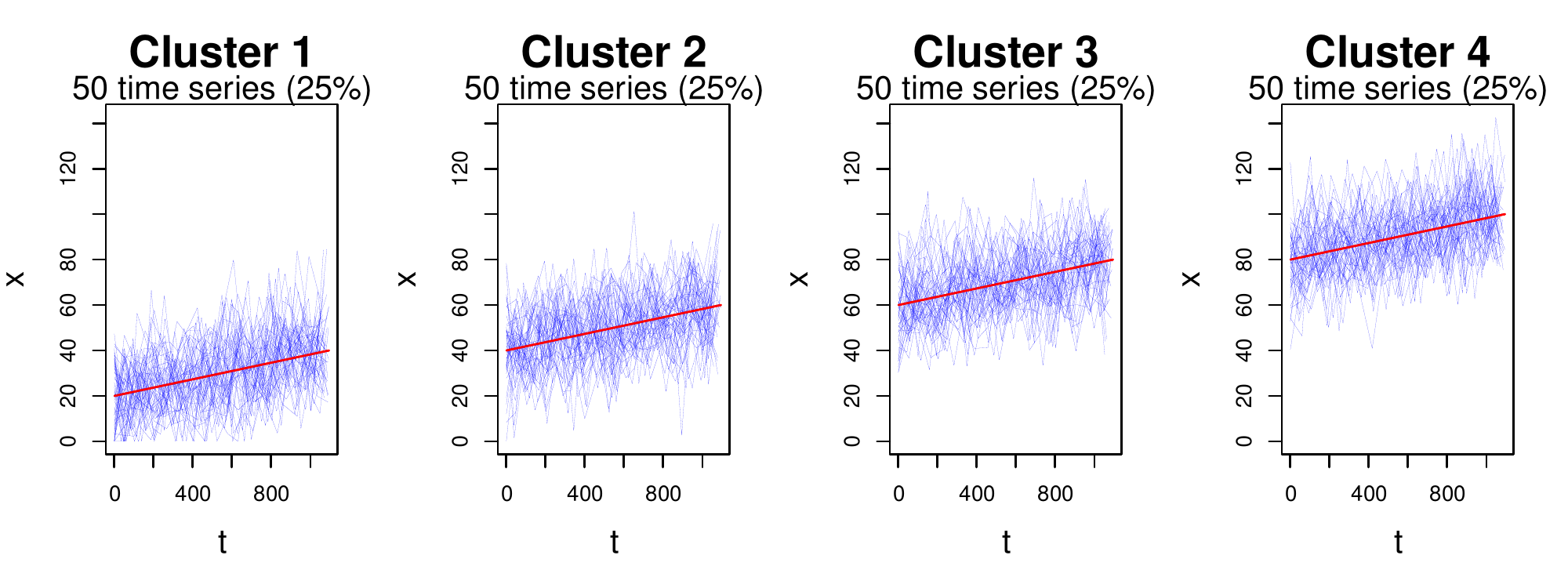}
		\caption{Synthetic dataset 2}
	\end{subfigure}
	\begin{subfigure}[c]{0.5\textwidth}
		\includegraphics[trim={0.45cm 0 0.45cm 0.3cm},clip, width=28mm, angle =-90]{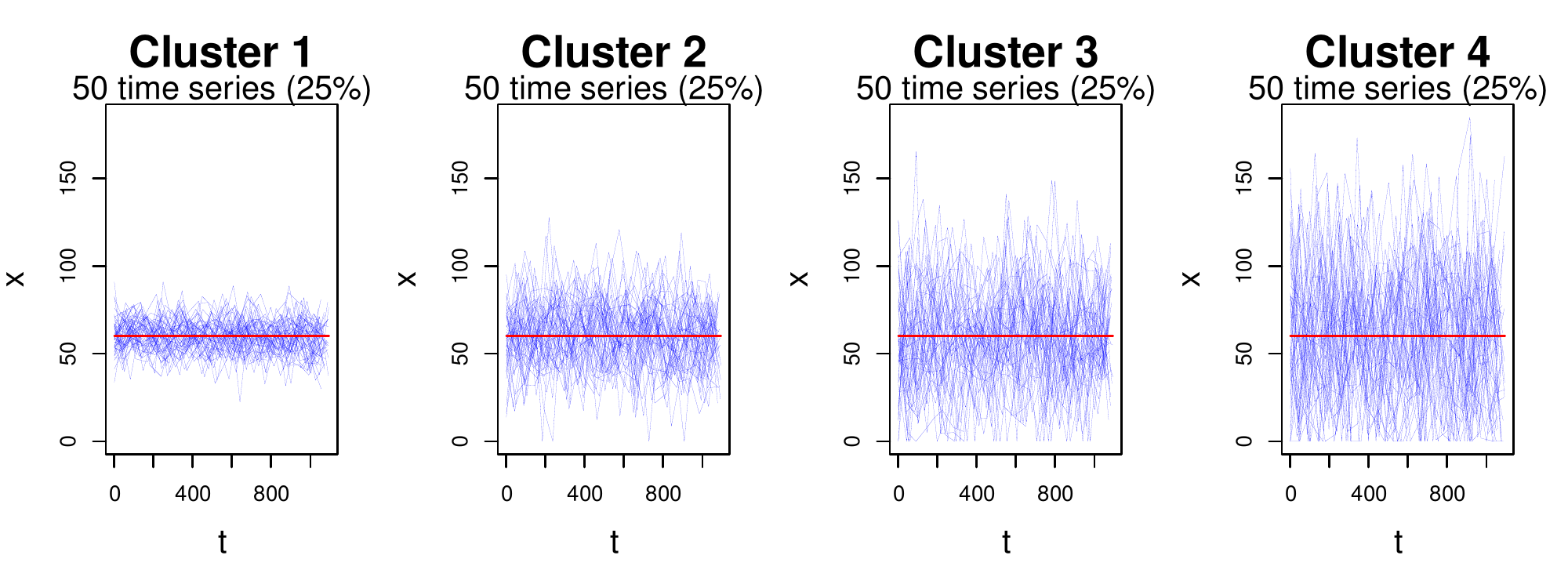}
		\caption{Synthetic dataset 3}
	\end{subfigure}
	\caption{Visualization of three synthetic datasets}
	\label{fig:synthetic_data}
\end{figure}

To simplify our analysis, we only consider longitudinal profile in which each observation has one real value. 
This is also similar to our application of analyzing CKD that disease progression is monitored via one primary clinical marker - eGFR that measures kidney function. 
In order to judge the quality of the embedded representations of longitudinal profiles, we will use three synthetic datasets that we know in prior the underlying mechanism of generating data to determine the role of hidden and memory unit in representing longitudinal profile. 
Each synthetic dataset consists of four clusters and each cluster contains $50$ longitudinal profiles. 
Each profile belonging to a cluster is sampled by first sampling the $t$ value that represents the time stamp an observation is made and the $x$ value is then sampled by following the underlying line that controls the behavior of the cluster as shown in Table~\ref{tab:synthetic_data}. 
The first sample of a profile is always at time $0$ and time gap for subsequent sample is computed by sampling from Poisson distribution with mean value $50$, i.e. $\Delta_t \sim Poisson(\lambda = 50)$. 
This means that the time gap between consecutive observations is approximately $50$ time steps. 
When time step exceeds $1095$, we stop sampling process for the profile. 
Figure~\ref{fig:synthetic_data} shows the visualization of our synthetic datasets. 
As can be observed in both the Figure~\ref{fig:synthetic_data} and the Table~\ref{tab:synthetic_data}, our synthetic datasets are prepared in a way that each cluster has only one difference from the other within the same dataset. 
For synthetic dataset 1, each line in the cluster has the same x-intercept and same level of variation and the only difference is the slope. 
For synthetic dataset 2, all lines within each cluster have the same slope and same level of variations while different clusters have different x-intercepts. 
For synthetic dataset 3, all clusters are basically horizontal lines with different levels of noises. 

\begin{table*}[!ht]
	\centering
	\begin{tabular}{ l | l | l | l }
		& Synthetic dataset 1 & Synthetic dataset 2 & Synthetic dataset 3 \\ 
		\hline
		Cluster 1 & $x = (4/219)t + 60 + \epsilon$ & $x = (4/219)t + 20 + \epsilon$ & $x = 60 + \epsilon_1$ where $\epsilon_1 \sim \mathcal{N}(0, 10)$ \\  
		Cluster 2 & $x = 60 + \epsilon$ & $x = (4/219)t + 40 + \epsilon$ & $x = 60 + \epsilon_2$ where $\epsilon_2 \sim \mathcal{N}(0, 20)$ \\
		Cluster 3 & $x = (-4/219)t + 60 + \epsilon$ & $x = (4/219)t + 60 + \epsilon$ & $x = 60 + \epsilon_2$ where $\epsilon_3 \sim \mathcal{N}(0, 30)$ \\
		Cluster 4 & $x = (-8/219)t + 60 + \epsilon$ & $x = (4/219)t + 80 + \epsilon$ & $x = 60 + \epsilon_2$ where $\epsilon_4 \sim \mathcal{N}(0, 40)$ \\
		& \multicolumn{2}{c}{where $\epsilon \sim \mathcal{N}(0, 14)$} &  \\
	\end{tabular}
	\caption{Summary of three synthetic datasets}
	\label{tab:synthetic_data}
\end{table*}

\begin{figure*}[!ht]
	\centering
	\begin{subfigure}[c]{0.32\textwidth}
		\centering
		\includegraphics[width=54mm, height=27mm]{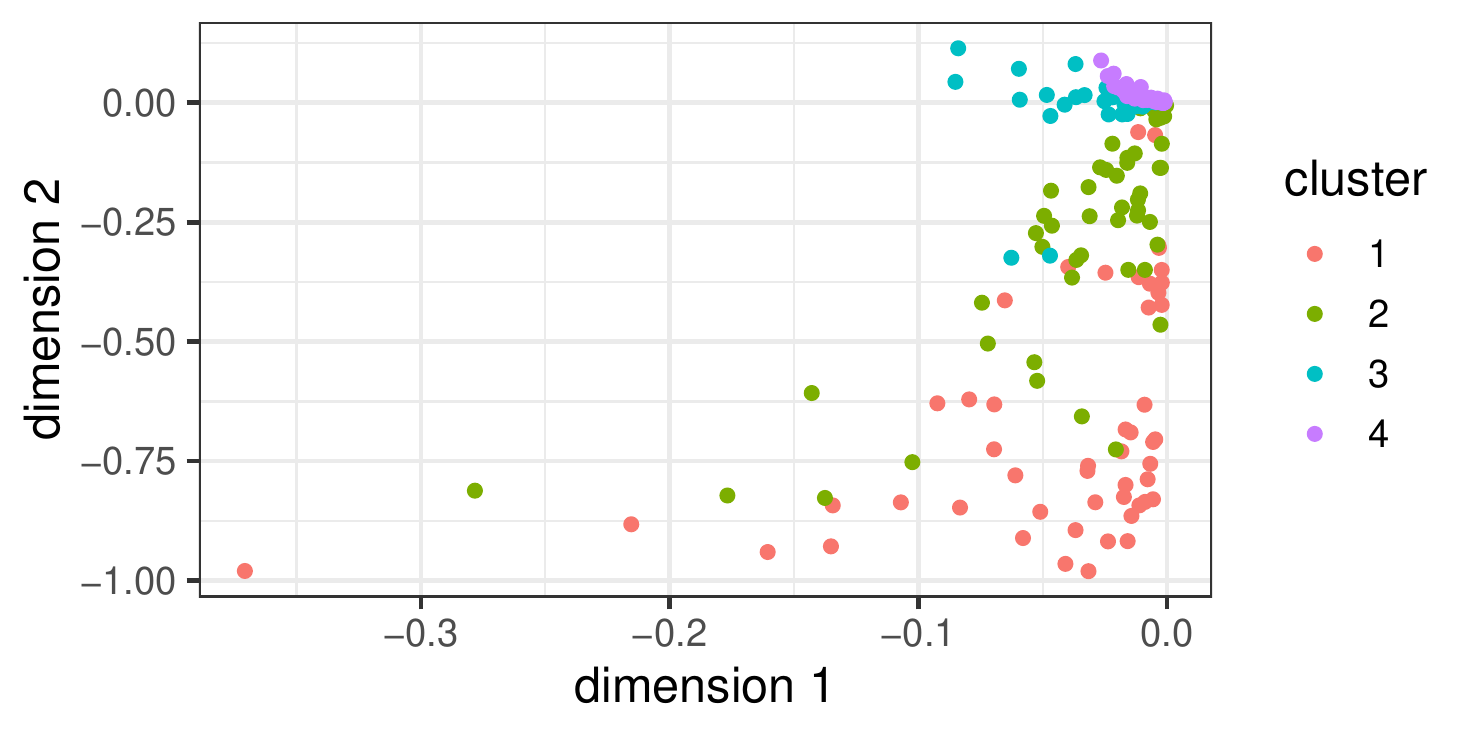}
		\caption{Hidden unit as longitudinal profile representations in synthetic dataset 1}
		\label{fig:data1_hidden}
	\end{subfigure}
	\begin{subfigure}[c]{0.32\textwidth}
		\centering
		\includegraphics[width=54mm, height=27mm]{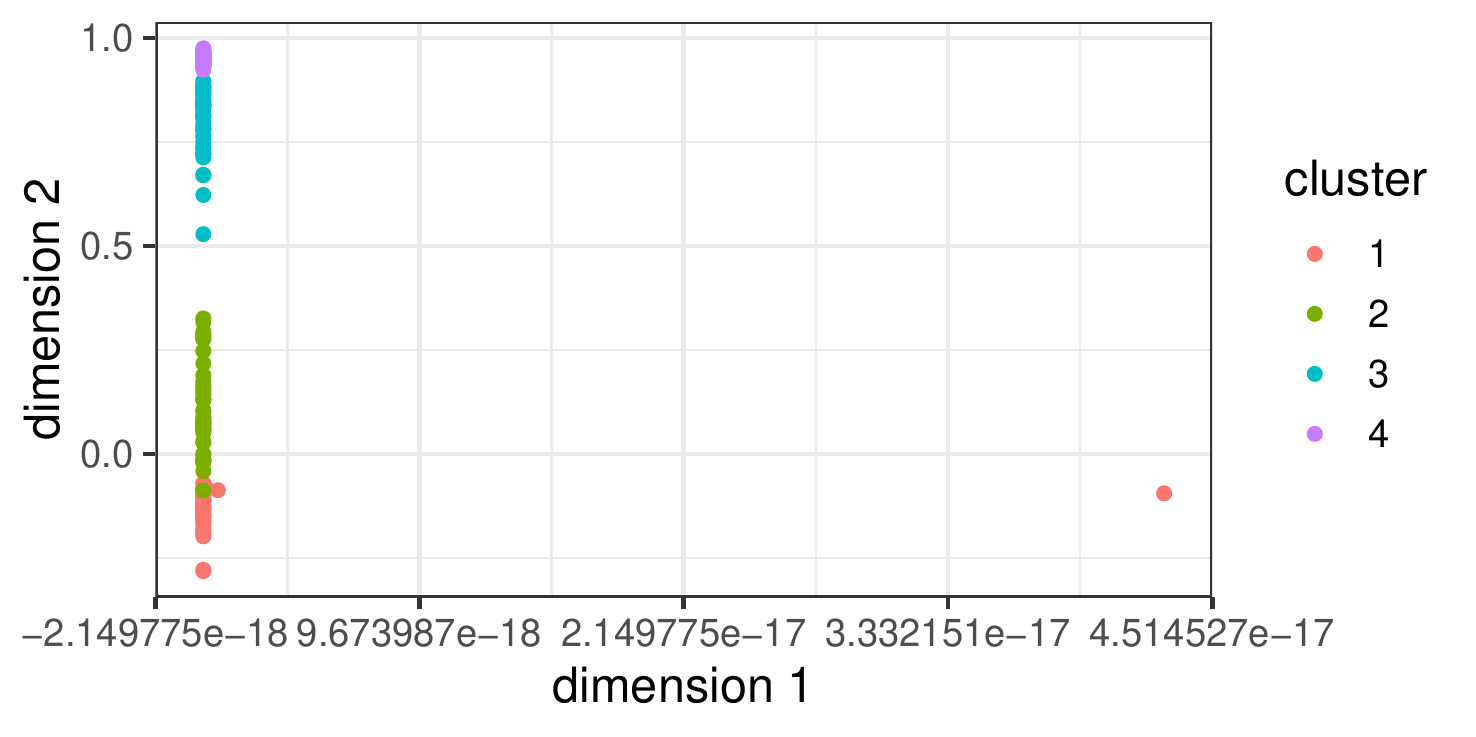}
		\caption{Hidden unit as longitudinal profile representations in synthetic dataset 2}
		\label{fig:data2_hidden}
	\end{subfigure}
	\begin{subfigure}[c]{0.32\textwidth}
		\centering
		\includegraphics[width=54mm, height=27mm]{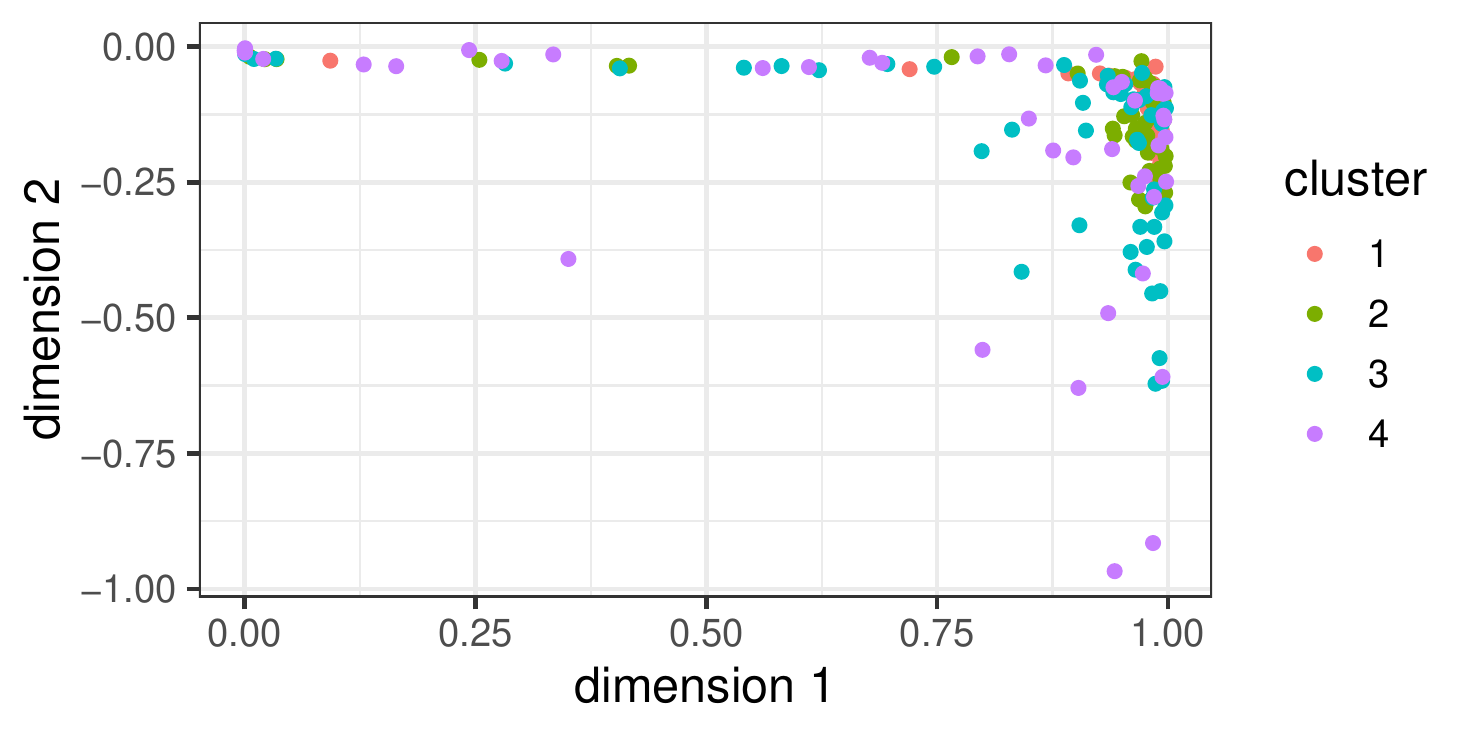}
		\caption{Hidden unit as longitudinal profile representations in synthetic dataset 3}
		\label{fig:data3_hidden}
	\end{subfigure}
	\begin{subfigure}[c]{0.33\textwidth}
		\centering
		\includegraphics[width=54mm, height=27mm]{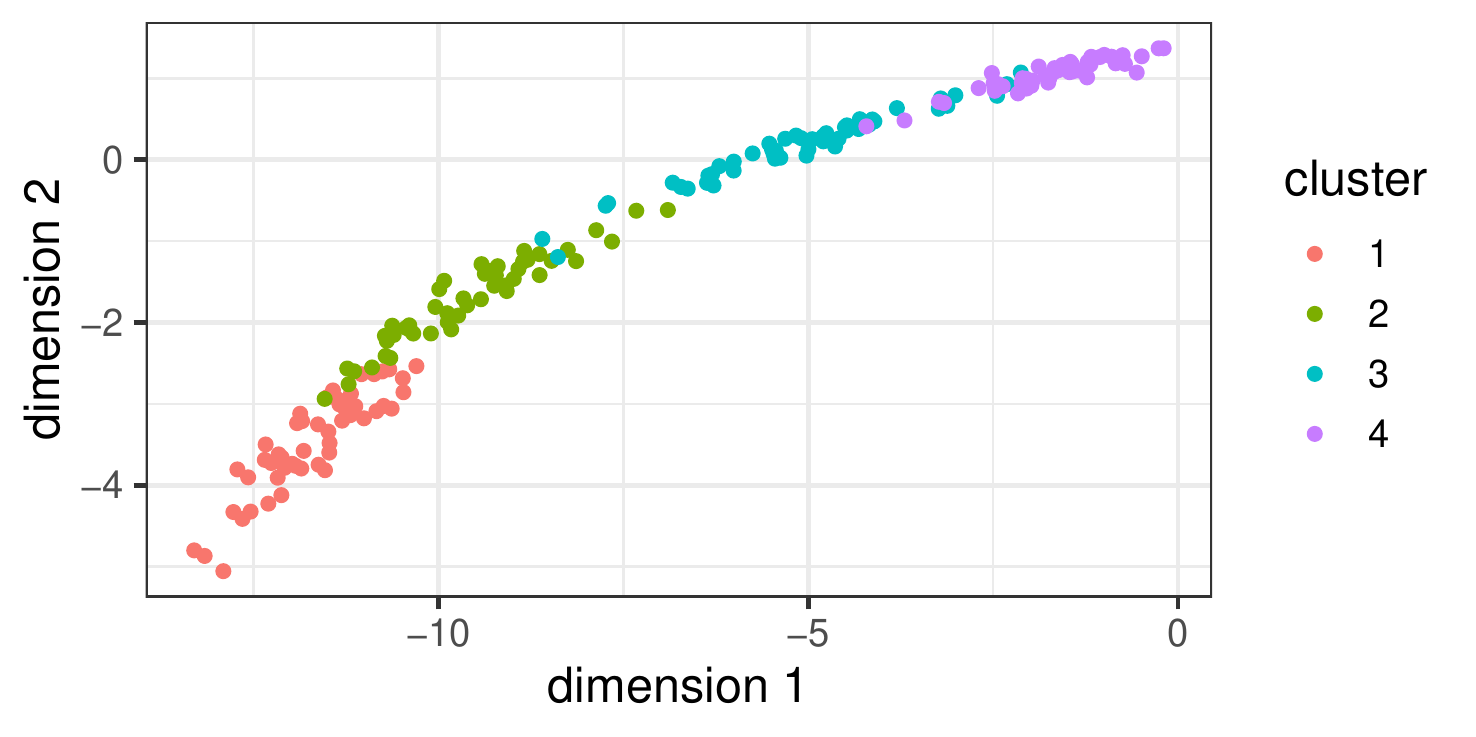}
		\caption{Memory unit as longitudinal profile representations in synthetic dataset 1}
		\label{fig:data1_mem}
	\end{subfigure}
	\begin{subfigure}[c]{0.32\textwidth}
		\centering
		\includegraphics[width=54mm, height=27mm]{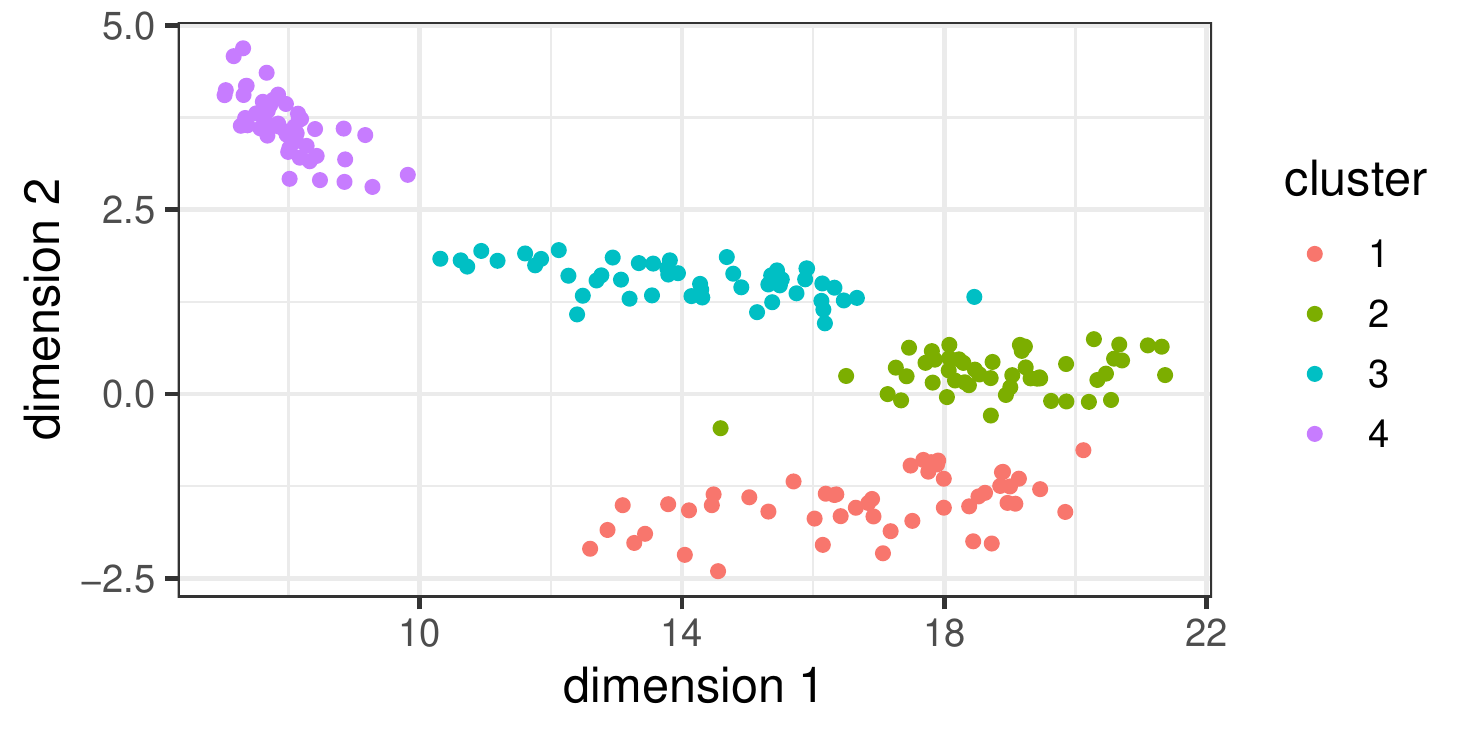}
		\caption{Memory unit as longitudinal profile representations in synthetic dataset 2}
		\label{fig:data2_mem}
	\end{subfigure}
	\begin{subfigure}[c]{0.32\textwidth}
		\centering
		\includegraphics[width=54mm, height=27mm]{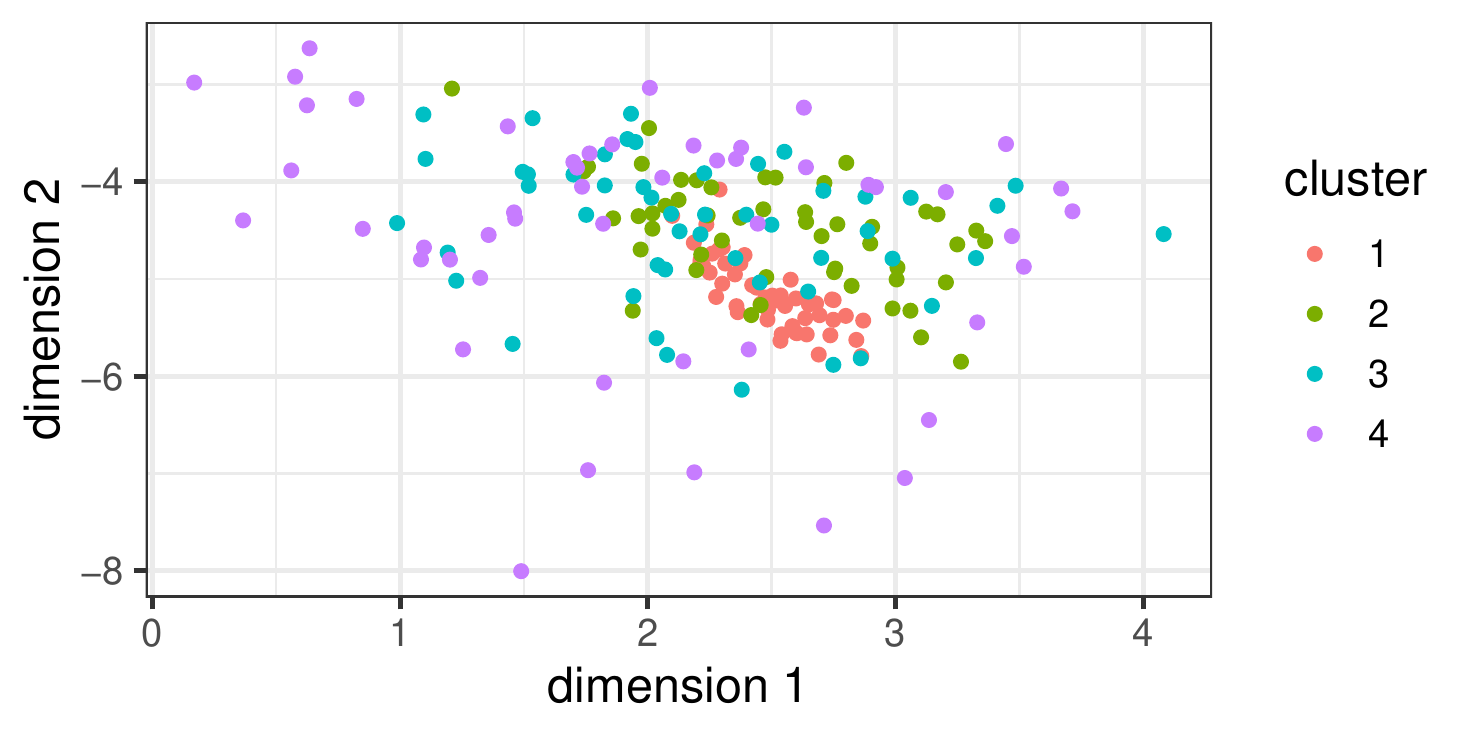}
		\caption{Memory unit as longitudinal profile representations in synthetic dataset 3}
		\label{fig:data3_mem}
	\end{subfigure}
	\caption{Visualization of hidden unit and memory unit at the last time step of encoder of T-LSTM Autoencoder after being trained for $20,000$ epochs (Best view in color)}
	\label{fig:tlstm_representation}
\end{figure*}

Because in all three synthetic datasets, there are only one distinctive difference between clusters  (slope for synthetic dataset 1, intercept for synthetic dataset 2 and noise level for synthetic dataset 3), the dimension of hidden unit (and also memory unit) is set to 2 as it is enough to capture the underlying differences between clusters, given that T-LSTM Autoencoder is able to capture this sole difference in the model. 
The two-dimensional hidden space in hidden unit (and also in memory unit) allows us to visualize the distribution of longitudinal profiles and inspect the alignment of the representations with their ground-truth clusters.
Figure~\ref{fig:tlstm_representation} shows visualization of the representations of individual longitudinal profiles in three datasets after being trained for $20,000$ epochs. 
This visualization provides some interesting insights regarding the role of hidden units and memory units at the last time step in the encoder as longitudinal profile representations.

As we can see in figure~\ref{fig:data1_hidden} and figure~\ref{fig:data1_mem}, the distribution of profile representations in synthetic dataset 1 does align with the ground-truth clusters, and this alignment is more visible in memory units than in hidden units. 
To quantify the alignment of representations in memory units / hidden units with the ground-truth clusters, we compute the silhouette coefficients~\cite{rousseeuw1987silhouettes} with ground-truth cluster assignments and distance metric between two longitudinal profiles is the Euclidean distance between their embedded representations. 
In Silhouette coefficient, the quality of individual data point in a clustering result with respect to a cluster assignment is quantified by the cluster tightness and degree of separation between neighboring clusters. 
The value of silhouette coefficient ranges from -1 to 1 in which higher value indicates better cluster assignment. 
The overall quality of a clustering result can be measured as the average of individual Silhouette coefficients. 
Table~\ref{tab:sil} shows the average silhouette coefficient when 3 types of longitudinal representations are used: (1) only hidden unit, (2) only memory unit and (3) both hidden and memory unit. 
As shown in Table~\ref{tab:sil}, for synthetic dataset 1, memory unit is significantly better in distinguishing four different clusters in comparison with hidden unit. 
When using both hidden unit and memory unit, the clustering quality is still better than hidden unit while only suffers minor reduction in clustering quality
in comparison with memory unit.

\begin{table}[!ht]
	\centering
	\begin{tabular}{ l | l | l  }
		& Synthetic  & Synthetic  \\ 
		Representation & dataset 1 & dataset 2\\
		\hline
		Hidden unit & $0.0548$ & $0.6169$  \\  
		Memory unit & $0.4906$ & $0.4769$  \\
		Hidden and memory unit & $0.4858$ & $0.4812$ \\
	\end{tabular}
	\caption{Average Silhouette coefficient when using ground-truth cluster membership}
	\label{tab:sil}
\end{table}

For synthetic dataset 2, we can observe in Figure~\ref{fig:data2_hidden} and~\ref{fig:data2_mem} that there are clear clustering patterns in the visualization of longitudinal representation by memory unit and hidden unit. 
It is worth to notice that for representation by hidden unit as shown in Figure~\ref{fig:data2_hidden}, the difference between clusters only concentrate in dimension 2 while values in dimension 1 are almost the same for each individual longitudinal profile. 
On the other hand, when looking at representation by memory unit as shown in Figure~\ref{fig:data2_mem}, there are clear four clusters spanning in both two dimensions of memory unit. 
When judging quantitatively the distinguishing power of hidden unit and memory unit, we also use the average silhouette coefficient as similar to synthetic dataset 1. 
As shown in Table~\ref{tab:sil}, for synthetic dataset 2, all three types of representations are well-aligned with the cluster ground-truth as their average silhouette coefficient is very high (the lowest value is $0.4769$ for the case of using memory unit as representation).
These average silhouette coefficients show that both hidden and memory unit are well-suited for representing the longitudinal profile, as both of them can capture well the differences in the x-intercept between clusters. 

For synthetic dataset 3, we can see in Figure~\ref{fig:data3_hidden} that representations by using hidden unit do not provide any clear clustering patterns. On the other hand, in Figure~\ref{fig:data3_mem}, 
most of data points of cluster 1 (cluster with lowest level of noise) concentrates in the center of the space while data points of cluster 4 (cluster with highest level of noise) scatter along the periphery of the space. For cluster 2 and 3, they are blended to some degree between center and the periphery of the space. This is an interesting observation as the level noise tends to play a role in determining the position of the longitudinal profile in latent space of memory unit. 

From the examination of the representations of longitudinal profiles in three different synthetic datasets, we can see that memory unit does play an important role in distinguishing different clusters. In a situation as in synthetic dataset 2, using hidden unit as representation does provide better distinguishing power between clusters. Therefore, in subsequent experiments with real clinical datasets, we use both hidden unit and memory unit as the representation for longitudinal profile.

\section{Experiments with Chronic Kidney Disease}
\label{sec:experiment}
In this section, we apply T-LSTM Autoencoder to explore the distribution of CKD longitudinal profiles in the latent space and investigate how the embedded representations provide better insights about the progression of CKD.
In section~\ref{subsec:data}, we briefly describe two CKD datasets that are used in the experiments.
In section~\ref{subsec:dimension}, we determine an appropriate dimension of hidden unit (and also memory unit) for T-LSTM Autoencoder to represent the embedded patient profile.
In section~\ref{subsec:reconstruction}, we investigate the reconstruction of original longitudinal profiles output from the decoder of T-LSTM Autoencoder and see how the inputs and outputs are related to each other.
In section~\ref{subsec:ckd}, we assess the embedded representations of longitudinal profiles learnt from T-LSTM Autoencoder and identify interesting and unusual longitudinal CKD progressions enabled by these latent representations.

\subsection{CKD datasets}
\label{subsec:data}
\begin{figure}[!ht]
	\centering
	\includegraphics[trim={7cm 0 7cm 0}, scale = 0.3]{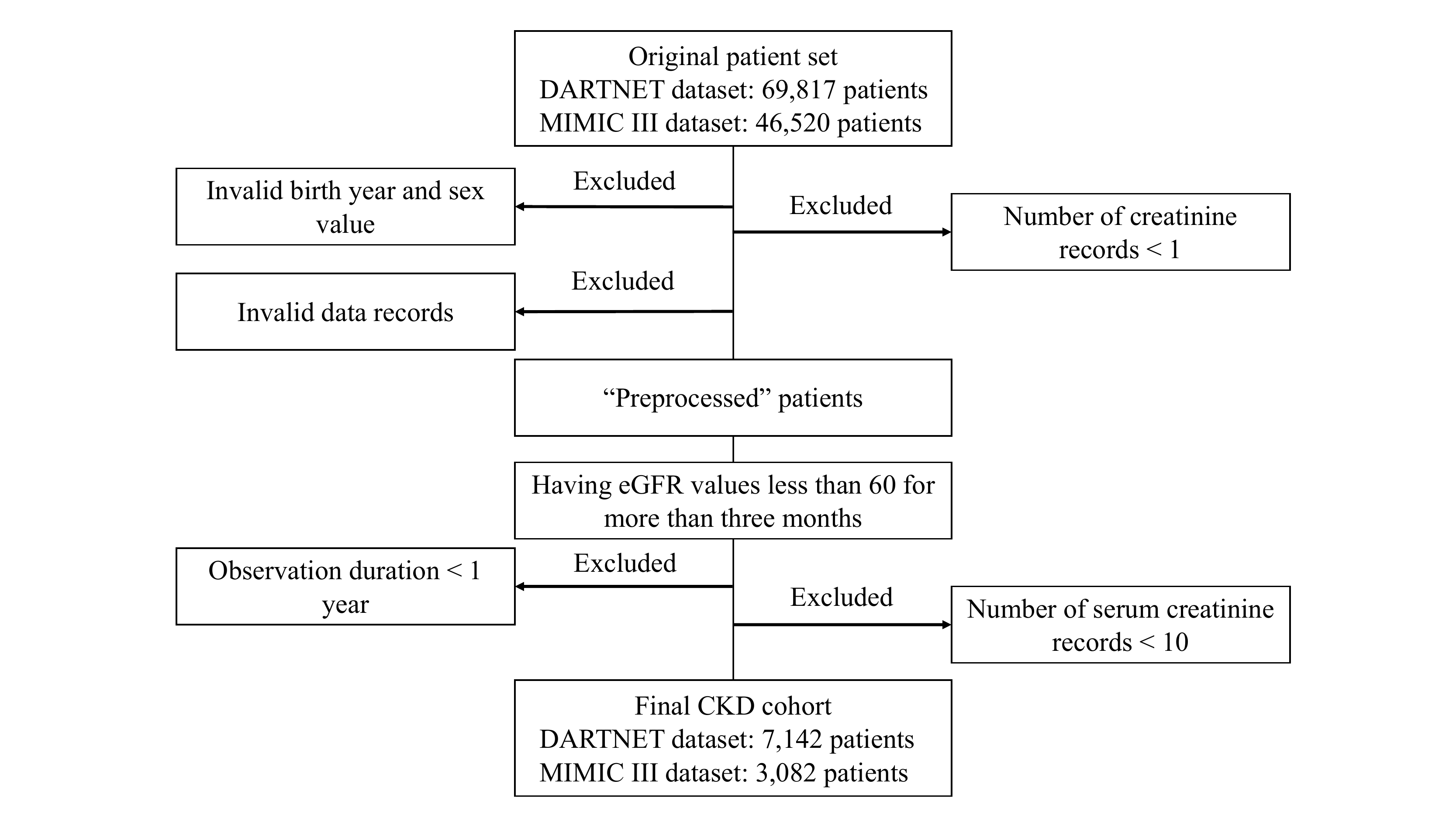}
	\caption{Preprocessing process to obtain CKD cohorts}
	\label{fig:preprocess}
\end{figure}

In our experiments, we use two real-world CKD datasets extracted from two larger clinical datasets: (1) DARTNet dataset~\cite{pace2014dartnet} and (2) MIMIC-III dataset~\cite{johnson2016mimic}. 
From the original datasets, we only consider patients having stage 3 CKD with more than ten eGFR observations and clinical records spanning more than one year to include in our experiments. 
This preprocessing step is similar to earlier study of Luong and Chandola~\cite{luong2017kmeans}. 
After preprocessing, we have $7,142$ patients and $3,082$ patients in CKD cohorts of DARTNet and MIMIC-III dataset respectively.
Figure~\ref{fig:preprocess} shows the preprocessing steps that we performed to obtain CKD cohorts, which is similar to earlier study of Luong and Chandola~\cite{luong2017kmeans}.
 
\subsection{Dimension of hidden units}
\label{subsec:dimension}
One of the key hyper-parameter of T-LSTM Autoencoder is the dimension of hidden unit (and also memory unit). 
This determines the representations of longitudinal profiles after being learned from T-LSTM Autoencoder. 
In earlier experiments with synthetic datasets, we have decided the dimension of hidden/memory unit to be two as it is enough to capture the variability in our synthetic datasets. 
However, in the real-world clinical datasets, there are many factors that can affect the progression of a longitudinal clinical profile. 
Therefore, choosing an appropriate dimension for our embedded representations is not a trivial task. 
In earlier study of Baytas et al.~\cite{baytas2017patient}, authors chose hidden dimension to be two for the case of synthetic dataset while they did not provide a clear criterion for choosing the hidden dimension when applying the T-LSTM Autoencoder on the real clinical dataset. 
In this section, we provide a rigorous process to determine dimension of hidden unit for clinical datasets.

As mentioned earlier, T-LSTM Autoencoder optimizes the reconstruction error. 
In other words, the reconstructed time series output from the decoder should be as close as possible to the original time series input to the encoder. 
Therefore, one criterion to estimate the quality of T-LSTM Autoencoder is via the reconstruction error. 
As T-LSTM Autoencoder optimizes this quantity directly using mini-batch Adam optimizer, the overall reconstruction error should decrease over many epochs during the training phase.
However, overfitting may occur as the T-LSTM Autoencoder may have very small reconstruction error for a training set but fail to provide good reconstruction for a new set of longitudinal profiles that have not been trained before.

For this reason, we use 4-fold cross-validation to determine the appropriate dimension of hidden/memory unit. 
In particular, the longitudinal profiles in each dataset are split into four parts. 
T-LSTM model is trained on each three parts (training set) while being validated on the remaining part (validation set).
The root-mean-square error (RMSE) between the reconstructed values and original values in the validation set is the measure for quality of T-LSTM Autoencoder.
The lower of RMSE in validation set, the better quality of T-LSTM Autoencoder is as it can generalize well to unseen data.
Finally, the overall RMSE is computed by using the RMSE computed for each fold in the cross-validation process, i.e. $\sqrt{\frac{RMSE_1^2 + \cdots + RMSE_4^2}{4}}$.

\begin{figure}[!ht]
	\centering
	\begin{subfigure}[c]{0.24\textwidth}
		\centering
		\includegraphics[scale=0.18]{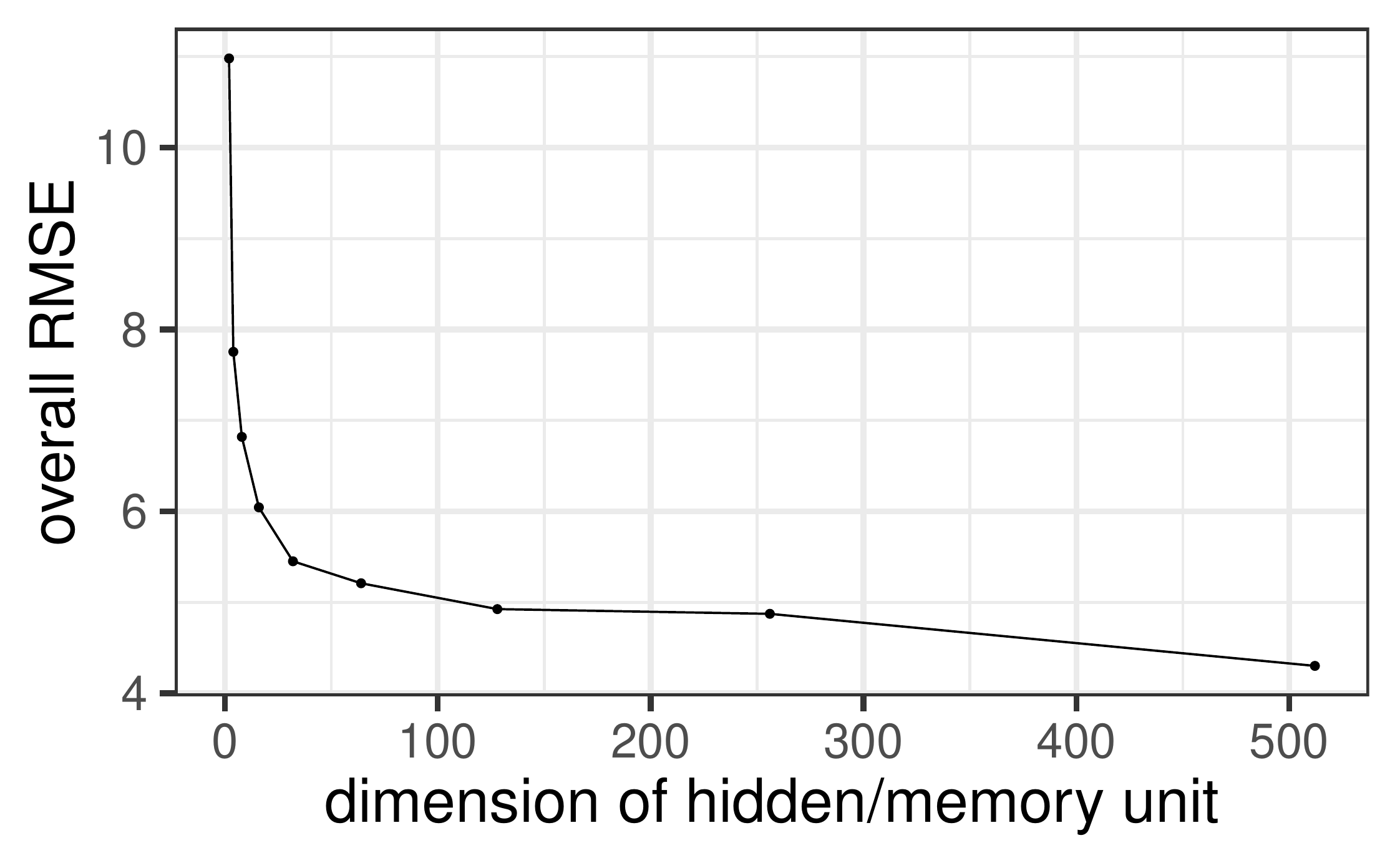}
		\caption{DARTNet dataset}
		\label{fig:cross_val_dartnet}
	\end{subfigure}
	\begin{subfigure}[c]{0.24\textwidth}
		\centering
		\includegraphics[scale=0.18]{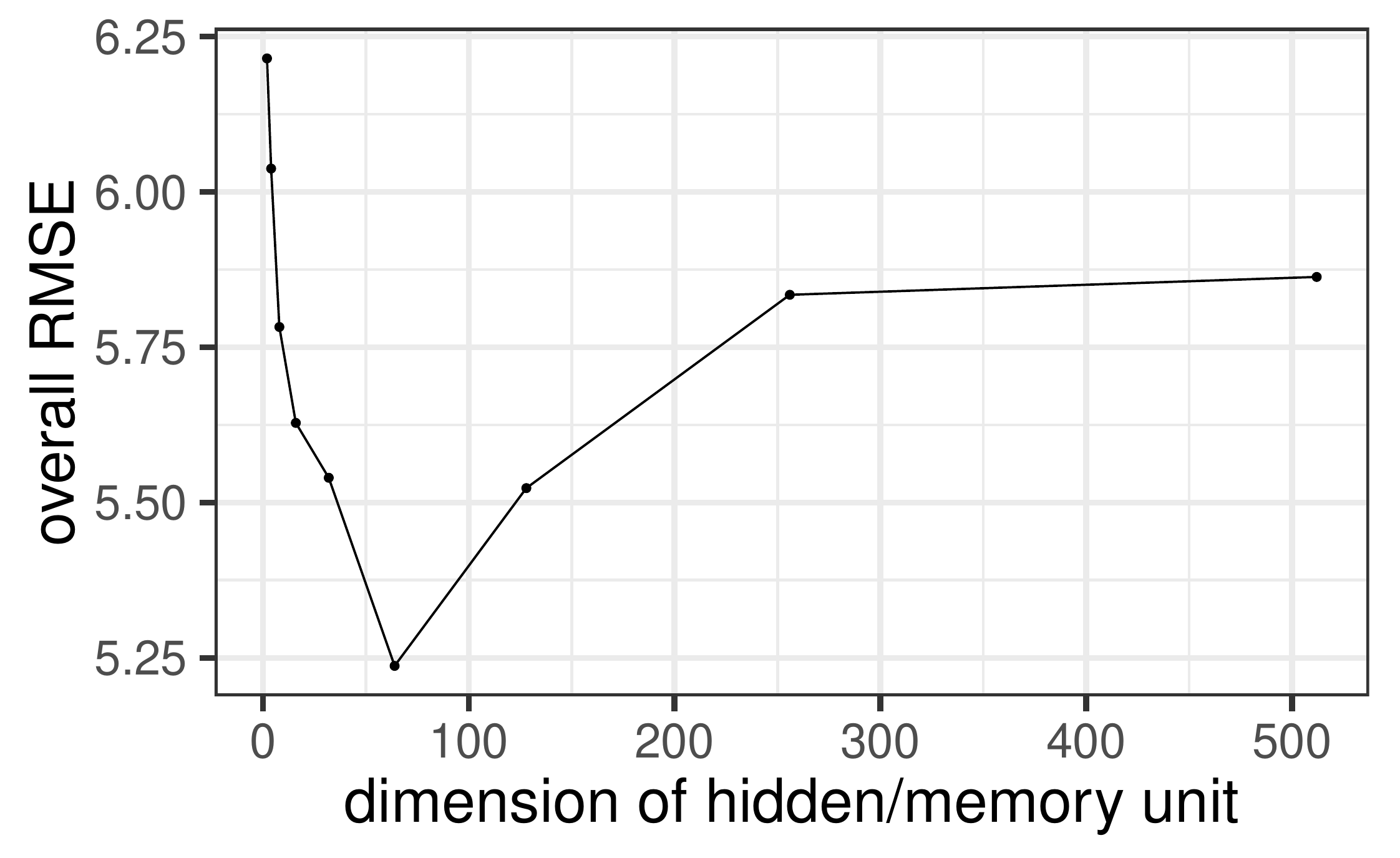}
		\caption{MIMIC-III dataset}
		\label{fig:cross_val_mimic}
	\end{subfigure}
	\caption{Overall RMSE of validation set with respect to different dimension of hidden/memory unit}
	\label{fig:cross_validation}
\end{figure}

We perform the above 4-fold cross validation on DARTNet and MIMIC-III datasets and vary the dimension of hidden/memory unit across following values $\{2^2, 2^3, 2^4, \cdots, 2^9\}$. 
In this experiment, we train T-LSTM Autoencoder with $500$ epochs.
Figure~\ref{fig:cross_validation} shows the overall RMSE of validation set with different values of hidden/memory dimension. 
For MIMIC-III dataset, we can observe in Figure~\ref{fig:cross_val_mimic} that $64$ is the optimal value for dimension of hidden/memory unit as the reconstruction error starts increasing after $64$, signaling that T-LSTM Autoencoder starts suffering from overfitting.
For DARTNet dataset, the overall RMSE keeps decreasing as we increase the dimension of hidden/memory unit, indicating that overfitting is not a problem in this situation.
However, as we can observe in Figure~\ref{fig:cross_val_dartnet}, the overall RMSE decreases rapidly until reaching dimension of value $64$ and becomes saturated for dimension values larger than $64$.
Based on the above observations, we choose $64$ as the dimension of hidden/memory unit when applying T-LSTM Autoencoder on two CKD datasets in subsequent experiments. 
As we discuss earlier in Section~\ref{sec:experiment_synthetic}, we will use both hidden and memory unit at the last time step of encoder to represent a longitudinal profile. 
Therefore, the dimension of our longitudinal profiles derived from T-LSTM Autoencoder is $128$ including $64$ dimensions from hidden unit and another $64$ dimensions from memory unit.

\subsection{Outputs of decoder}
\label{subsec:reconstruction}
\begin{figure*}[!ht]
	\centering
	\begin{subfigure}[c]{0.32\textwidth}
		\centering
		\includegraphics[scale=0.3]{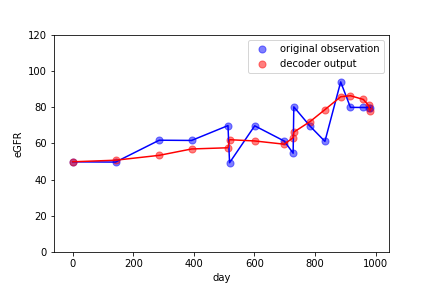}
		\caption{Patient profile with ID $*965$ in DARTNet dataset}
		\label{fig:output_decoder_dartnet1}
	\end{subfigure}
	\begin{subfigure}[c]{0.32\textwidth}
		\centering
		\includegraphics[scale=0.3]{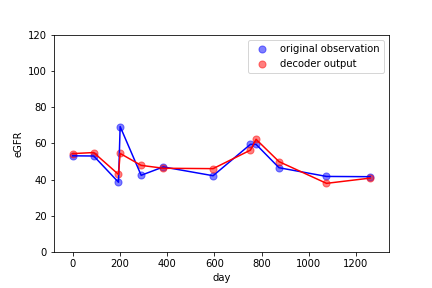}
		\caption{Patient profile with ID $*698$ in DARTNet dataset}
		\label{fig:output_decoder_dartnet2}
	\end{subfigure}
	\begin{subfigure}[c]{0.32\textwidth}
		\centering
		\includegraphics[scale=0.3]{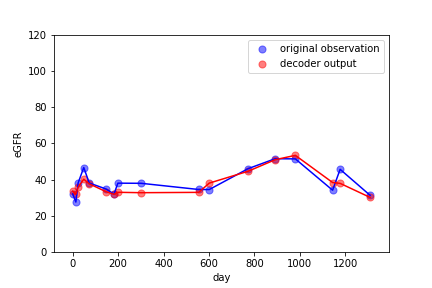}
		\caption{Patient profile with ID $*838$ in DARTNet dataset}
		\label{fig:output_decoder_dartnet3}
	\end{subfigure}
	\begin{subfigure}[c]{0.32\textwidth}
		\centering
		\includegraphics[scale=0.3]{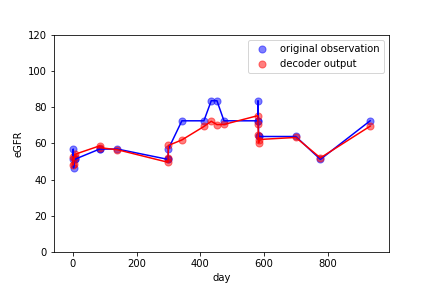}
		\caption{Patient profile with ID $*03$ in MIMIC-III dataset}
		\label{fig:output_decoder_mimic1}
	\end{subfigure}	
	\begin{subfigure}[c]{0.32\textwidth}
		\centering
		\includegraphics[scale=0.3]{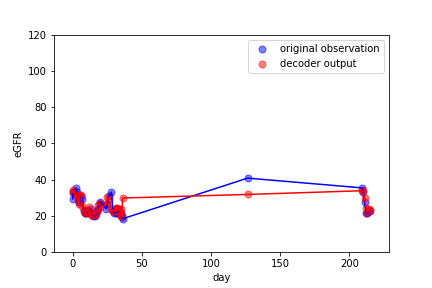}
		\caption{Patient profile with ID $*91$ in MIMIC-III dataset}
		\label{fig:output_decoder_mimic2}
	\end{subfigure}
	\begin{subfigure}[c]{0.32\textwidth}
		\centering
		\includegraphics[scale=0.3]{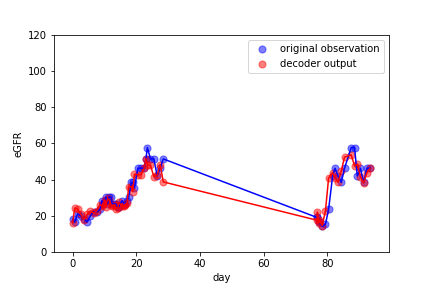}
		\caption{Patient profile with ID $*70$ in MIMIC-III dataset}
		\label{fig:output_decoder_mimic3}
	\end{subfigure}
	\caption{Reconstructed time series in comparison with original time series. (Best view in color)}
	\label{fig:decoder}
\end{figure*}

As we described earlier, the outputs of decoder in T-LSTM Autoencoder are actually the prediction of the original values in the longitudinal profile in reverse chronological order (see Figure~\ref{fig:tlstm_ae} and Section~\ref{subsec:autoencoder}). 
One of the expectations when applying T-LSTM Autoencoder into CKD is that the disease progression reconstructed from the model should be able to capture the same progression of the original longitudinal profile. 
In this section, we investigate the reconstructed time series obtained from outputs of the decoder and observe how similar they are to the original time series. 
Figure~\ref{fig:decoder} shows the outputs of the decoder in comparison with the original longitudinal data for three particular patients in each dataset.
In the figure, the reconstructed time series from outputs of decoder can capture the long-term progression in patient CKD profile.
In addition, for some particular time stamps in which the original CKD profile experiences sudden jumps in eGFR value, we can observe that the reconstructed time series have smoother transitions from one time step to another time step.
In other words, we can view output of decoder in T-LSTM Autoencoder as a denoising summarizer that can capture the trends in CKD progression while simplifying the time series by reducing short-term variance.

In order to verify this observation for the whole dataset, we perform a statistical test to see whether the variance of the reconstructed time series is significantly smaller than the variance original time series. 
We first compute the variance in original time series and in reconstructed time series for each patient profile. 
After that, we use the pairs of variances, one for each patient in the dataset to perform a t-test to test whether there is a significant larger amount of variance from the original time series in comparison with the reconstructed one. 
The p-values of this t-test are both less than $2.2\times 10^{-16}$ for both DARTNet and MIMIC-III dataset. 
This shows that the variance of reconstructed time series is significantly less than variance in the original time series, which confirms our observation.

\subsection{Embedded representation}
\label{subsec:ckd}
\begin{figure*}[!ht]
	\centering
	\begin{subfigure}[l]{0.32\textwidth}
		\centering
		\includegraphics[scale=0.25]{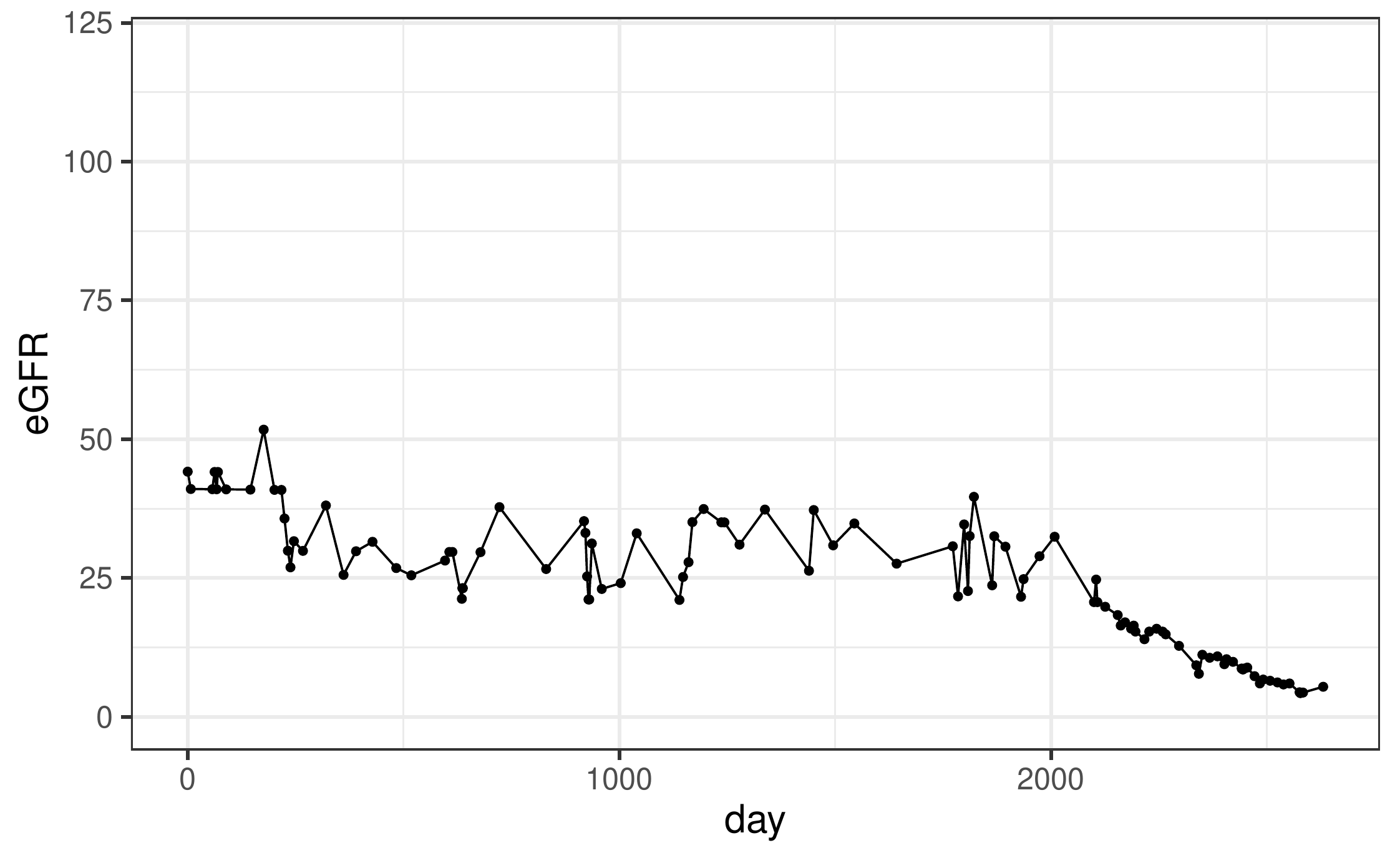}
		\caption{Patient $*476$ with extreme values in $15$ different dimensions}
		\label{fig:dartnet_outlier1}
	\end{subfigure}
	\centering
	\begin{subfigure}[c]{0.32\textwidth}
		\centering
		\includegraphics[scale=0.25]{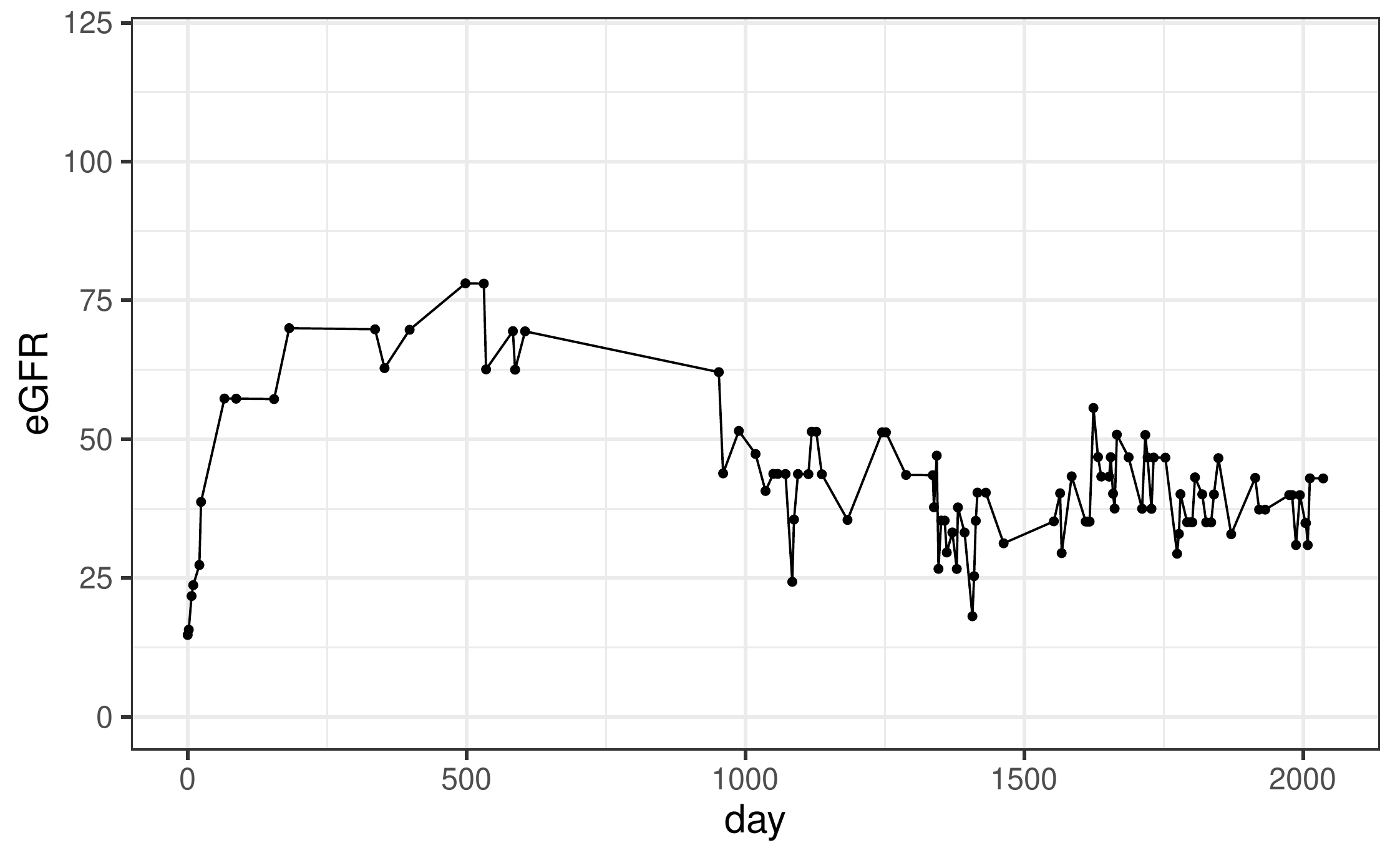}
		\caption{Patient $*598$ with extreme values in $11$ different dimensions}
		\label{fig:dartnet_outlier2}
	\end{subfigure}
	\centering
	\begin{subfigure}[r]{0.32\textwidth}
		\centering
		\includegraphics[scale=0.25]{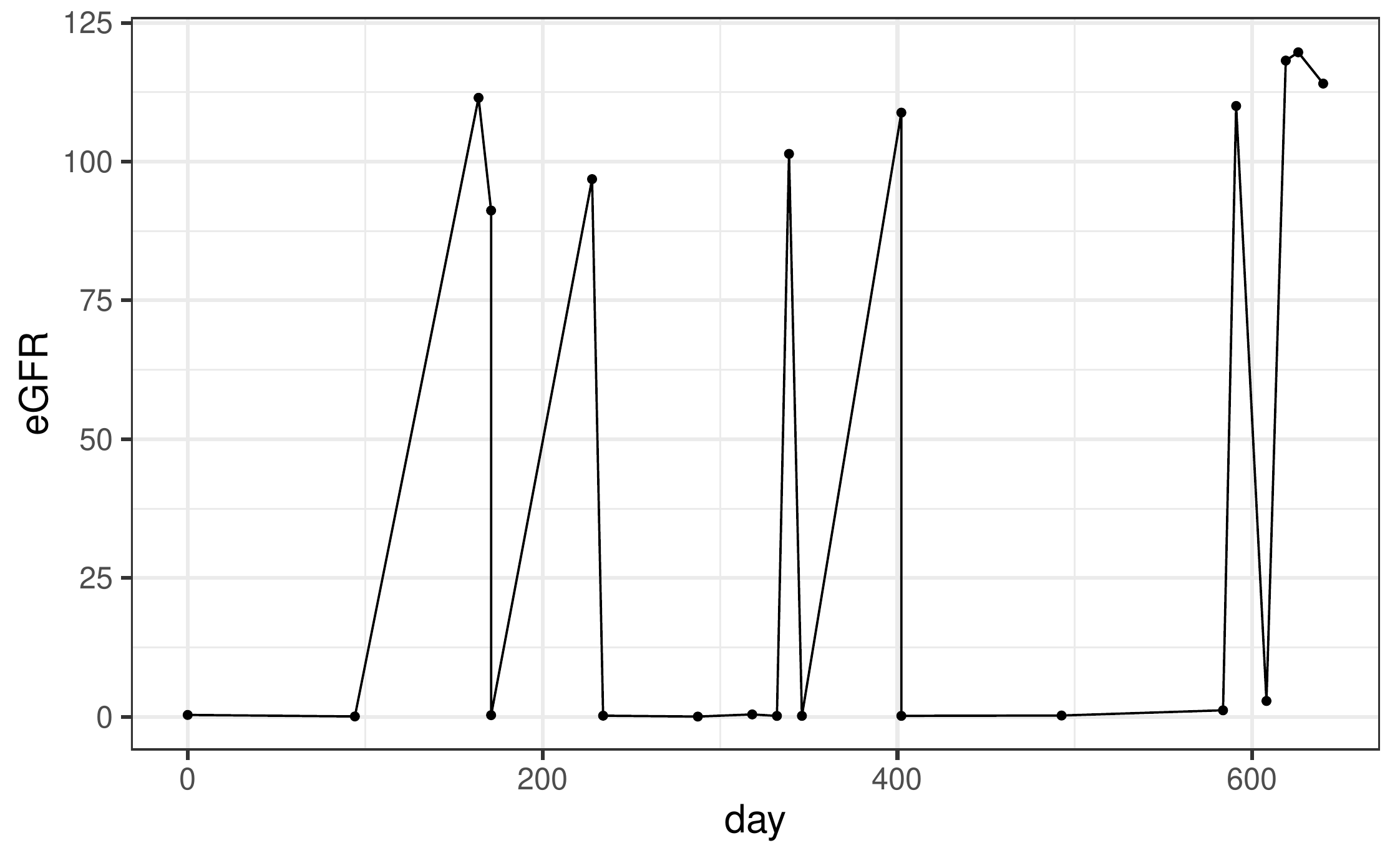}
		\caption{Patient $*534$ with extreme values in $7$ different dimensions}
		\label{fig:dartnet_outlier3}
	\end{subfigure}
	\centering
	\begin{subfigure}[l]{0.32\textwidth}
		\centering
		\includegraphics[scale=0.25]{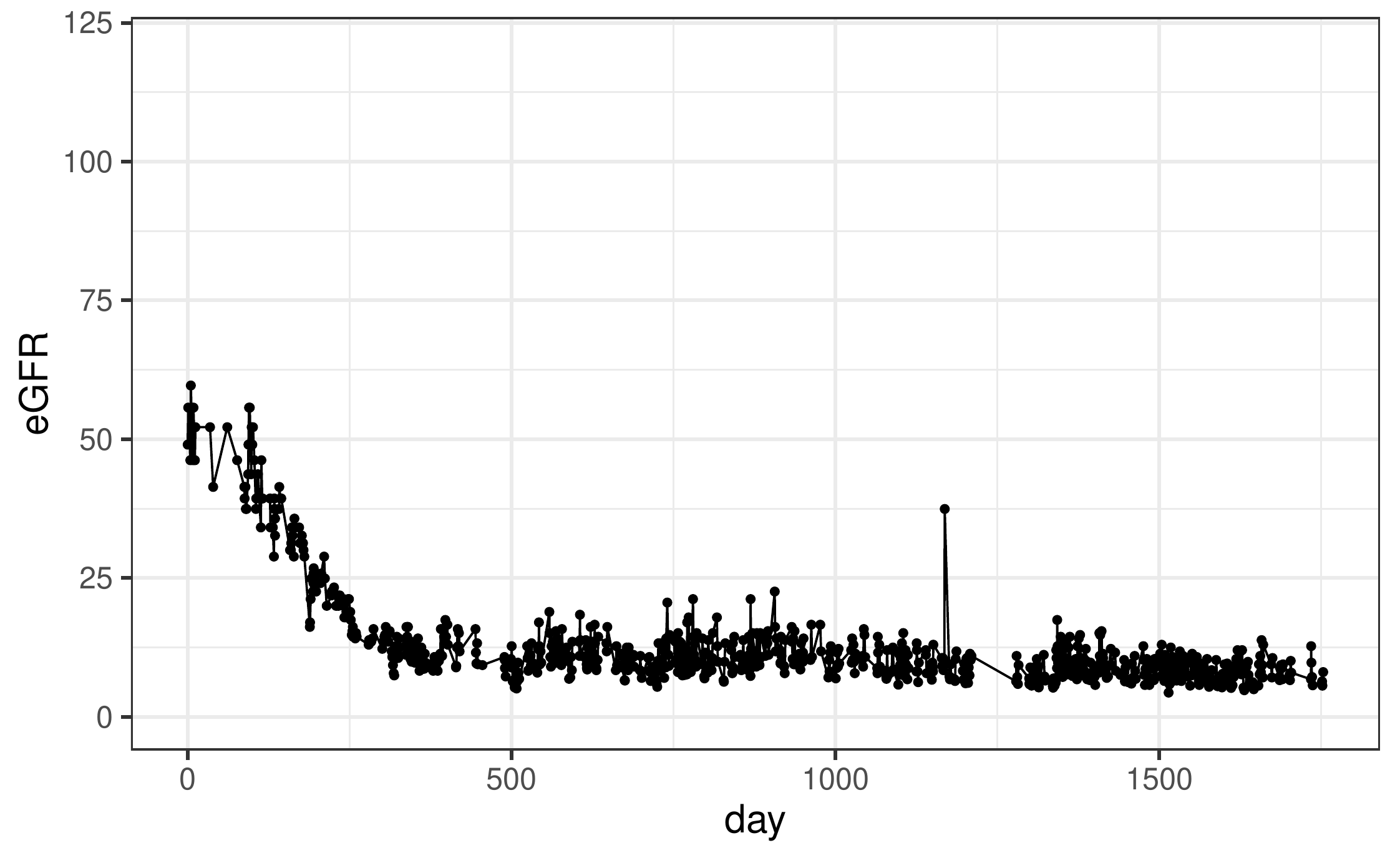}
		\caption{Patient $*33$ with extreme values in $16$ different dimensions}
		\label{fig:mimic_outlier1}
	\end{subfigure}
	\centering
	\begin{subfigure}[c]{0.32\textwidth}
		\centering
		\includegraphics[scale=0.25]{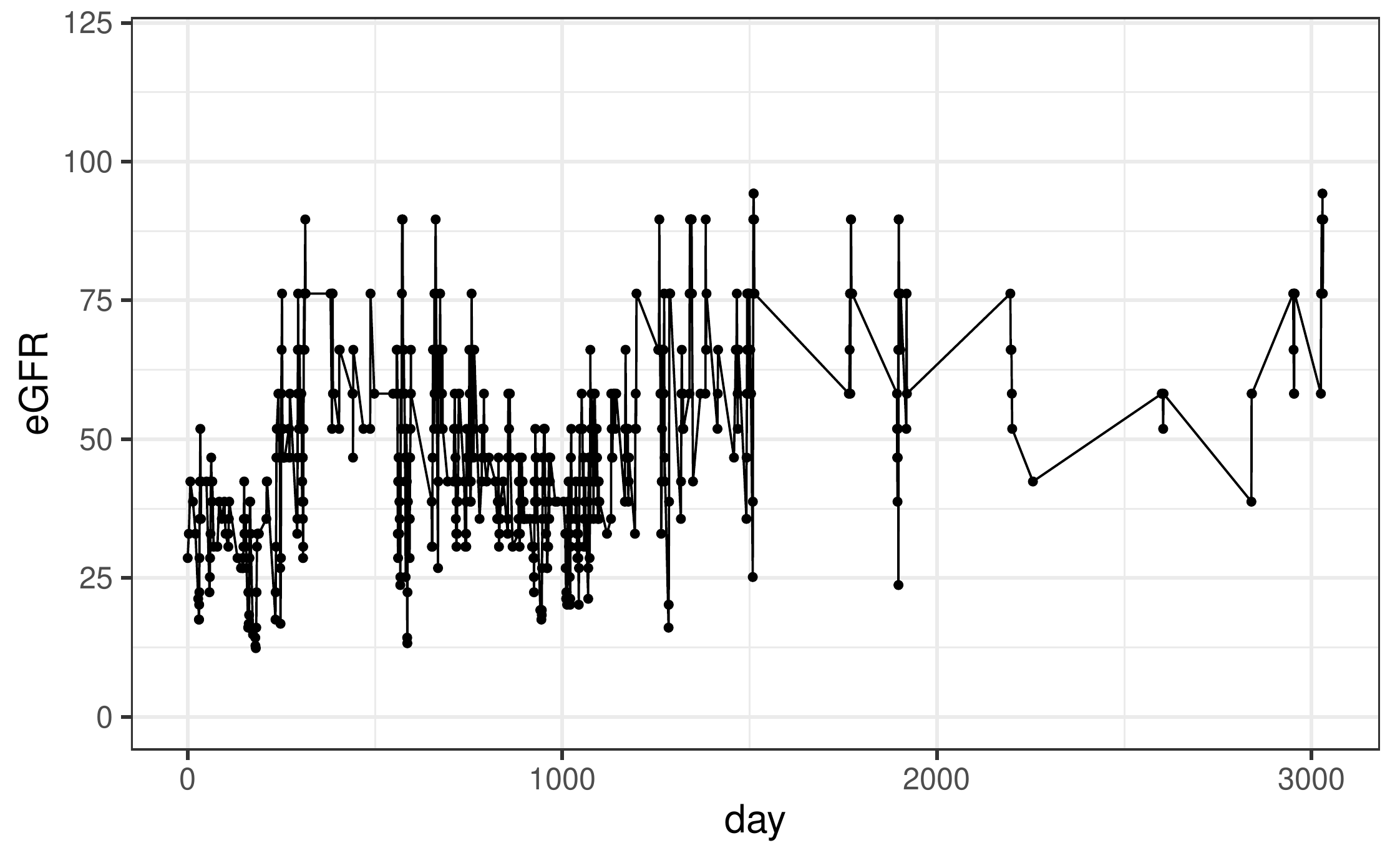}
		\caption{Patient $*18$ with extreme values in $15$ different dimensions}
		\label{fig:mimic_outlier2}
	\end{subfigure}
	\centering
	\begin{subfigure}[r]{0.32\textwidth}
		\centering
		\includegraphics[scale=0.25]{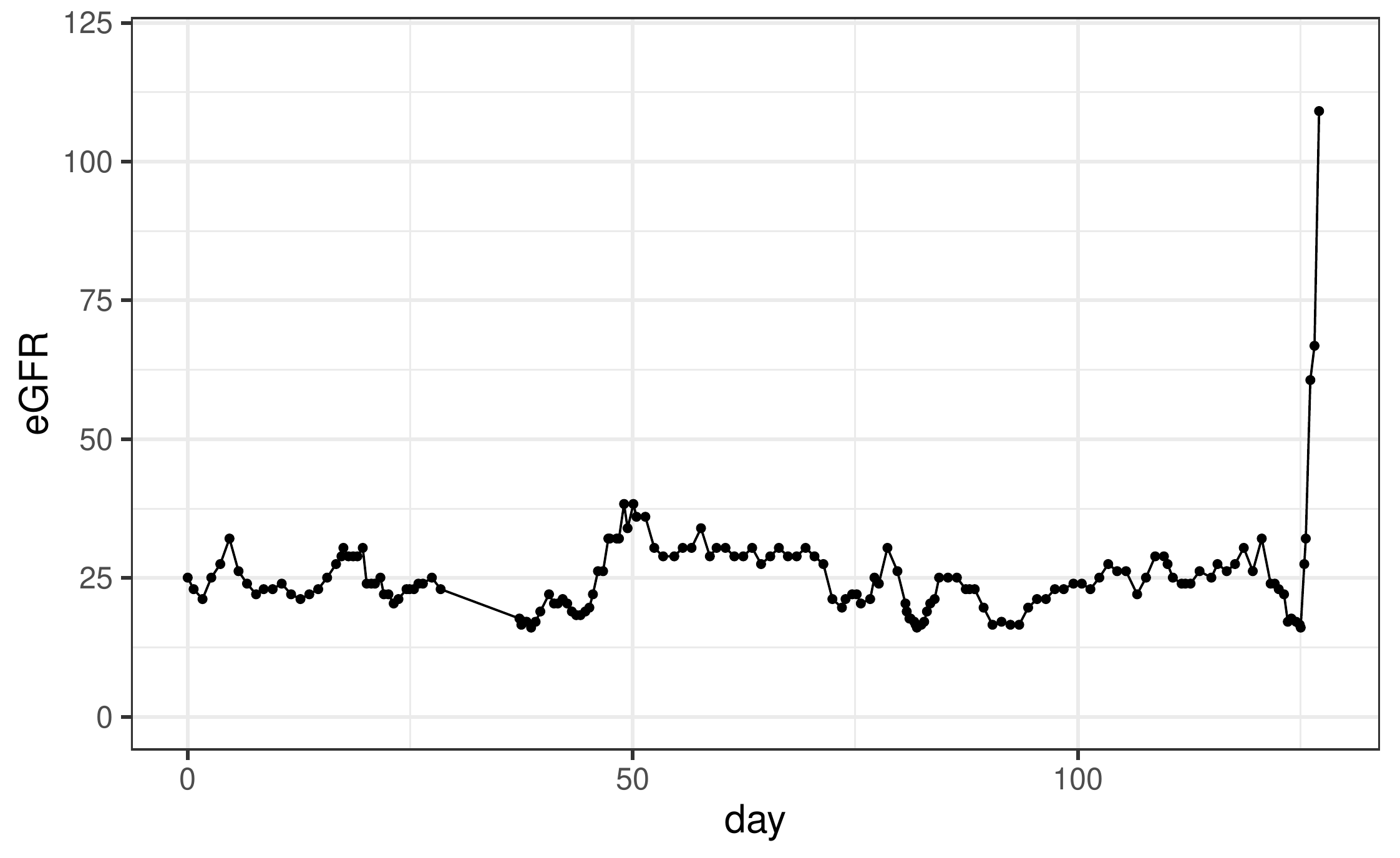}
		\caption{Patient $*14$ with extreme values in $7$ different dimensions}
		\label{fig:mimic_outlier3}
	\end{subfigure}
	\caption{Individual CKD profiles which embedded representations have extreme values in multiple dimensions. Patients in Figure (a)-(c) are in DARTNet dataset. Patients in Figure (d)-(f) are in MIMIC-III dataset.}
	\label{fig:interesting_profiles}
\end{figure*}

In this section, we investigate the embedded patient profiles obtained by T-LSTM Autoencoder. 
As described in Section~\ref{subsec:dimension}, each patient profile is actually represented by a $128$ dimensional vector. 
When analyzing the progression in CKD, it is important to detect patients with interesting and unusual disease progressions. 
In other words, this task can be viewed as detecting outlier time series from a dataset of many irregularly sampled time series. 
By applying T-LSTM Autoencoder, each of the longitudinal profile has been transformed into a vector of fixed dimension. 
Each dimension in this embedded representation is an encoding of complex temporal features of the original longitudinal profile. 
One way to extract the outlier from this set of longitudinal profiles is to look at each individual dimension and check for the value that is farthest from the mean value of this dimension.
In our experiment, by following this approach, we have found a set of few longitudinal profiles that they contain extreme values (farthest from the mean of a particular dimension) in multiple dimensions.
Figure~\ref{fig:interesting_profiles} shows the set of individual profiles that contain ``outliers" in multiple different dimensions. 
From this figure, we can identify some of the uniqueness in those longitudinal profiles. 

First observation we have in Figure~\ref{fig:interesting_profiles} is that extreme length of the sequence can cause the embedded representation to have extreme values. In fact, patient $*476$ and $*598$ have two longest sequences in DARTNet dataset with $107$ and $102$ eGFR readings respectively. Similarly, patient $*33$ and $*18$ are two longest sequences in MIMIC-III dataset with $877$ and $655$ eGFR readings respectively.\footnote{On average, a patient profile in DARTNet dataset has only $15.8$ eGFR observations while a patient in MIMIC-III dataset has only $49.8$ observations.} 

The high fluctuation in consecutive eGFR readings may also be a contributing factor to cause the extreme values. In particular, patient profiles in Figure~\ref{fig:dartnet_outlier3} is highly unusual in the sense that eGFR readings fluctuate in a very wide range between $0$ (severe kidney damage) and $100$ (healthy kidney function). This can be a result of incorrect reading values by mistakes of input software when entering inputs to the EHR system. We can also observe the fluctuation in Figure~\ref{fig:mimic_outlier2} of patient $*18$. This patient profile has highly fluctuated observations over a long period of time. However, the range of fluctuation for the observations of this patient is not as extreme as previous patient. Therefore, we suspect that the measurement error instead of input error is the cause of this fluctuated observations.

Another interesting profile is of patient $*14$ in Figure~\ref{fig:mimic_outlier3}. This patient has a stable eGFR readings over the course of more than 100 days and then unexpectedly experiences a sudden increase in eGFR to more than $100$ in the last few days of his/her records. This is an interesting case as his/her course of disease is different from our understanding of CKD disease progression. Normally, kidney function in CKD patient deteriorates over the time and as a result, eGFR values will decrease over the time. However, this patient experiences a sudden improvement in eGFR readings, to a level of normal person with healthy kidney function, which is contrary to normal CKD disease progression. Therefore, this patient is a good candidate to further study his/her course of disease progression, which can subsequently help us expand our knowledge about CKD.

\section{Discussion and Related Works}
\label{sec:discussion}
One thing to note in T-LSTM Autoencoder is that its training time is highly dependent on the characteristics of training dataset as each step of the optimization requires the training time series to have the same length.
For this reason, we have to split the training data into many batches, and each batch contains the set of time series that have the same length.
This causes the training process to be dependent on the training dataset. 
In particular, for a dataset with many sequences having the same length, the training process will be efficient if it can take advantage of parallel computation when training multiple sequences together in each batch.
On the other hand, for a dataset with many sequences having very different lengths (which is true for many longitudinal clinical datasets), the training process is reduced to a similar form of stochastic gradient-based optimization as most of the batches contain only one sequence. 

%
In principle, T-LSTM Autoencoder can be used for longitudinal data with multi-dimensional observation. 
In our analysis on CKD, there is only one dimensional observation being used: eGFR - a main clinical marker of CKD. 
One may extend the model to have multi-dimensional observation by including other health indicators such as blood pressures, measures of liver functions, measures of body cholesterol level, etc. 
However, such an extended model may have to deal with missing values in the observation. 
For example, a patient who has observation of eGFR on a particular day may not have blood pressures and other clinical markers measured on the same day. 
In a study of Che et al.~\cite{che2018recurrent}, authors proposed a modification of GRU architecture - another RNN variation that has been showed to perform comparably with LSTM - to handle missing values in clinical observations.
However, the focus of Che et al.'s study is supervised learning in which it attempts to predict future events.
In our future works, we will further study the capability of T-LSTM Autoencoder to represent missing values in the observation vector.

Med2Vec~\cite{choi2016multi} and Wave2Vec~\cite{yuan2018wave2vec} are two earlier studies that attempt to learn the representation of clinical observations by using unsupervised learning. 
In particular, both Med2Vec and Wave2Vec are inspired by Skip-gram model~\cite{mikolov2013distributed}.
In the context of Natural Language Processing, the latent representations of words are learnt in a way that maximizes their co-occurrences with their nearby words. 
For clinical longitudinal dataset, both Med2Vec and Wave2Vec consider that the sequence of hospital visits for each patient is analogous to the sequence of words in a sentence in which we can learn the embedded representation of a hospital visit or a patient clinical observation by maximizing the co-occurrence of the current observation with few other observations prior to or after the current one.
However, instead of providing the embedded representations for the entire sequence of clinical observations like in the case of T-LSTM Autoencoder, Med2Vec and Wave2Vec only give an embedded representation for each clinical observation in the sequence. 

In another study of Schulam and Arora~\cite{schulam2016disease}, authors attempted to map irregularly sampled clinical time series into latent space by assuming patient's clinical observations follow Gaussian Processes. 
This study of Schulam and Arora focused on understanding whether there are a small number of degrees of freedom (dimensions of the latent space) that can explain the differences in the progressions of patient. 
In principle, T-LSTM Autoencoder can also be used to map the whole sequence of clinical observations into a small dimensional latent space, but at the expense of the higher reconstruction error.

\section{Conclusion}
\label{sec:conclusion}
In this paper, we have discussed various aspects of T-LSTM Autoencoder as a model to project longitudinal profiles into a latent space and its use on analyzing disease progression in CKD. 
With synthetic datasets, we demonstrated that both the memory and hidden unit at the last time step of the encoder should be used as the representation of longitudinal profile.
In real-world CKD datasets, we presented an approach of using cross-validation to determine the dimension of hidden/memory unit in T-LSTM Autoencoder. 
In addition, we observed that the outputs of the decoder after being trained on CKD datasets are generally smoother in comparison with the corresponding inputs, while they still capture the trends expressed in the original inputs. 
Finally, we used the embedded representations of patient eGFR longitudinal profiles learnt from T-LSTM Autoencoder to identify unusual and interesting CKD profiles.
The longitudinal profiles obtained by this experiment are candidates for two further analyses: (1) validate whether their inputs are correctly entered or not, and (2) in case the inputs are correctly entered, what is the underlying mechanism that causes the abnormality in the observations and whether we can enhance our understanding of CKD by understanding these unusual profiles.

\section*{Acknowledgment}
This material is based upon work supported by the National Science Foundation under award numbers CNS - 1409551 and IIS - 1641475.

\bibliographystyle{IEEEtran}
\bibliography{paper}{}

\begin{thebibliography}{10}
\providecommand{\url}[1]{#1}
\csname url@samestyle\endcsname
\providecommand{\newblock}{\relax}
\providecommand{\bibinfo}[2]{#2}
\providecommand{\BIBentrySTDinterwordspacing}{\spaceskip=0pt\relax}
\providecommand{\BIBentryALTinterwordstretchfactor}{4}
\providecommand{\BIBentryALTinterwordspacing}{\spaceskip=\fontdimen2\font plus
\BIBentryALTinterwordstretchfactor\fontdimen3\font minus
  \fontdimen4\font\relax}
\providecommand{\BIBforeignlanguage}[2]{{%
\expandafter\ifx\csname l@#1\endcsname\relax
\typeout{** WARNING: IEEEtran.bst: No hyphenation pattern has been}%
\typeout{** loaded for the language `#1'. Using the pattern for}%
\typeout{** the default language instead.}%
\else
\language=\csname l@#1\endcsname
\fi
#2}}
\providecommand{\BIBdecl}{\relax}
\BIBdecl

\bibitem{abubakar2015global}
I.~Abubakar, T.~Tillmann, and A.~Banerjee, ``Global, regional, and national
  age-sex specific all-cause and cause-specific mortality for 240 causes of
  death, 1990-2013: a systematic analysis for the global burden of disease
  study 2013,'' \emph{Lancet}, vol. 385, no. 9963, pp. 117--171, 2015.

\bibitem{luong2017egems}
D.~T.~A. Luong \emph{et~al.}, ``Extracting deep phenotypes for chronic kidney
  disease using electronic health records,'' \emph{eGEMs}, vol.~5, 2017.

\bibitem{luong2017kmeans}
D.~T.~A. Luong and V.~Chandola, ``A k-means approach to clustering disease
  progressions,'' in \emph{IEEE ICHI}, Aug 2017, pp. 268--274.

\bibitem{singh2017automatic}
P.~Singh, V.~Chandola, and C.~Fox, ``Automatic extraction of deep phenotypes
  for precision medicine in chronic kidney disease,'' in \emph{ICDH}, 2017, pp.
  195--199.

\bibitem{baytas2017patient}
I.~M. Baytas, C.~Xiao, X.~Zhang, F.~Wang, A.~K. Jain, and J.~Zhou, ``Patient
  subtyping via {Time-Aware} {LSTM} networks,'' in \emph{KDD}.\hskip 1em plus
  0.5em minus 0.4em\relax ACM, 2017, pp. 65--74.

\bibitem{WU2018167}
S.~Wu, S.~Liu, S.~Sohn, S.~Moon, C.~il~Wi, Y.~Juhn, and H.~Liu, ``Modeling
  asynchronous event sequences with {RNN}s,'' \emph{Journal of Biomedical
  Informatics}, vol.~83, pp. 167 -- 177, 2018.

\bibitem{hochreiter1997long}
S.~Hochreiter and J.~Schmidhuber, ``Long short-term memory,'' \emph{Neural
  computation}, vol.~9, no.~8, pp. 1735--1780, 1997.

\bibitem{rousseeuw1987silhouettes}
P.~J. Rousseeuw, ``Silhouettes: a graphical aid to the interpretation and
  validation of cluster analysis,'' \emph{Journal of computational and applied
  mathematics}, vol.~20, pp. 53--65, 1987.

\bibitem{pace2014dartnet}
W.~D. Pace, C.~Fox, T.~White, D.~Graham, L.~M. Schilling, and R.~David, ``The
  {DARTNet} institute: seeking a sustainable support mechanism for electronic
  data enabled research networks,'' \emph{eGEMs}, vol.~2, no.~2, p.~6, 2014.

\bibitem{johnson2016mimic}
A.~E. Johnson, T.~J. Pollard, L.~Shen, H.~L. Li-wei, M.~Feng, M.~Ghassemi,
  B.~Moody, P.~Szolovits, L.~A. Celi, and R.~G. Mark, ``{MIMIC-III}, a freely
  accessible critical care database,'' \emph{Scientific data}, vol.~3, 2016.

\bibitem{che2018recurrent}
Z.~Che, S.~Purushotham, K.~Cho, D.~Sontag, and Y.~Liu, ``Recurrent neural
  networks for multivariate time series with missing values,'' \emph{Scientific
  reports}, vol.~8, no.~1, p. 6085, 2018.

\bibitem{choi2016multi}
E.~Choi, M.~T. Bahadori, E.~Searles, C.~Coffey, M.~Thompson, J.~Bost,
  J.~Tejedor-Sojo, and J.~Sun, ``Multi-layer representation learning for
  medical concepts,'' in \emph{KDD}.\hskip 1em plus 0.5em minus 0.4em\relax
  ACM, 2016, pp. 1495--1504.

\bibitem{yuan2018wave2vec}
Y.~Yuan, G.~Xun, Q.~Suo, K.~Jia, and A.~Zhang, ``Wave2vec: Deep representation
  learning for clinical temporal data,'' \emph{Neurocomputing}, 2018.

\bibitem{mikolov2013distributed}
T.~Mikolov, I.~Sutskever, K.~Chen, G.~S. Corrado, and J.~Dean, ``Distributed
  representations of words and phrases and their compositionality,'' in
  \emph{NIPS}, 2013, pp. 3111--3119.

\bibitem{schulam2016disease}
P.~Schulam and R.~Arora, ``Disease trajectory maps,'' in \emph{NIPS}, 2016, pp.
  4709--4717.

\end{thebibliography}

\end{document}